\documentclass[lettersize,journal]{IEEEtran}
\usepackage{amsmath,amsfonts}
\usepackage{algorithmic}
\usepackage{algorithm}
\usepackage{array}
\usepackage[caption=false,font=normalsize,labelfont=sf,textfont=sf]{subfig}
\usepackage{textcomp}
\usepackage{stfloats}
\usepackage{url}
\usepackage{verbatim}
\usepackage{graphicx}
\usepackage{cite}
\usepackage{multirow}
\usepackage{hyperref}
\usepackage{xcolor}
\usepackage{tikz}

\begin{document}

\title{UniVRSE: Unified Vision-conditioned Response Semantic Entropy for Hallucination Detection in Medical Vision-Language Models}
\author{
Zehui Liao, 
Shishuai Hu,
Ke Zou,
Mengyuan Jin,
Yanning Zhang,
Huazhu Fu,
Liangli Zhen,
and Yong Xia
\thanks{
This work was supported in part by the National Natural Science Foundation of China under Grants 62171377 and 92470101, the `Pioneer' and `Leading Goose' R\&D Program of Zhejiang, China, under Grant 2025C01201(SD2), the Shenzhen Science and Technology Program under Grant JCYJ20220530161616036, the Ningbo Clinical Research Center for Medical Imaging under Grant 2021L003 (Open Project 2022LYKFZD06), the Innovation Foundation for Doctor Dissertation of Northwestern Polytechnical University under Grants CX2022056 and CX2023016, and the National Research Foundation Singapore under the National Cybersecurity R\&D Programme (Grand Challenge Award: CRPO-GC1-NTU-002).
(Corresponding author: Y. Xia, L. Zhen and H. Fu.)
}
\thanks{
Z. Liao, S. Hu, M. Jin, Y. Zhang and Y. Xia are with the National Engineering Laboratory for Integrated Aero-Space-Ground-Ocean Big Data Application Technology, School of Computer Science and Engineering, Northwestern Polytechnical University, Xi'an 710072, China (e-mail: merrical@mail.nwpu.edu.cn; sshu@mail.nwpu.edu.cn; myjin@mail.nwpu.edu.cn; ynzhang@nwpu.edu.cn; yxia@nwpu.edu.cn). 
K. Zou is with the National University of Singapore, Singapore 119077, Singapore. (e-mail: kezou18@163.com).
H. Fu and L. Zhen are with the Institute of High Performance Computing, Agency for Science, Technology and Research, Singapore 138632, Singapore. (e-mail: hzfu@ieee.org; zhenll@a-star.edu.sg).
}
}

\markboth{Journal of \LaTeX\ Class Files,~Vol.~14, No.~8, August~2021}%
{Liao \MakeLowercase{\textit{et al.}}: Unified Vision-conditioned Response Semantic Entropy for Hallucination Detection in Medical Vision-language Models}

\maketitle

\begin{abstract} 
Vision-language models (VLMs) hold significant promise for medical image understanding, particularly in open-ended tasks such as Visual Report Generation (VRG) and Visual Question Answering (VQA). 
However, they may produce hallucinations, referring to responses that conflict with visual evidence and pose a critical barrier to clinical deployment.
Although uncertainty-based hallucination detection methods are generally intuitive and effective, they remain limited in medical VLMs. In particular, Semantic Entropy (SE), which works well in text-only LLMs, becomes less reliable when applied directly to medical VLMs due to their overconfidence driven by strong language priors.
To address this challenge, we propose \textbf{UniVRSE}, a \textbf{Uni}fied \textbf{V}ision-conditioned \textbf{R}esponse \textbf{S}emantic \textbf{E}ntropy framework for hallucination detection in open-ended medical text generation. 
UniVRSE strengthens visual guidance during uncertainty estimation by contrasting the semantic predictive distributions derived from an original image–text pair and a visually distorted counterpart. 
Higher entropy in this discrepancy distribution reliably indicates hallucination risk. 
For VQA, UniVRSE operates directly on the image–question pair, whereas for VRG it decomposes the generated report into atomic claims, generates a verification question for each claim, and applies vision-conditioned entropy estimation at the claim level.
To evaluate hallucination detection, we construct a unified pipeline that generates responses on existing medical datasets and derives hallucination labels via factual consistency assessment using hallucination evaluation methods. 
However, current evaluation methods rely on subjective criteria or modality-specific rules. 
To improve reliability, we introduce Alignment Ratio of Atomic Facts (ALFA), a novel method that quantifies fine-grained factual consistency between the generated outputs and reference answers. 
ALFA-derived labels serve as ground truth for robust benchmarking hallucination detection.
Comprehensive experiments on six medical VQA/VRG datasets and three representative VLMs demonstrate that UniVRSE substantially surpasses existing detection methods and exhibits strong cross-modal generalization. 
The code will be released upon acceptance.
\end{abstract}


\section{Introduction}
\label{sec:intro}

\IEEEPARstart{M}{edical} vision–language models (VLMs)~\cite{xiao2024medvlmsurvey} have recently emerged as promising tools for open-ended clinical reasoning tasks, including visual report generation (VRG)~\cite{hartsock2024vision} and visual question answering (VQA)~\cite{liu2024gemex} across radiology, pathology, and ophthalmology. 
Trained via large-scale multi-modal pretraining, these models hold potential to improve diagnostic efficiency and clinical decision-making~\cite{messina2022survey}.
However, they still suffer from \textit{hallucinations}, referring to responses that are semantically plausible yet conflict with visual evidence, thus posing risks of misdiagnosis and diminished clinical trust~\cite{huang2023llmhallusurvey,liu2024mllmhallusurvey}.
Although recent efforts in data curation~\cite{yu2024hallucidoctor}, training regularization~\cite{yang2025mitigating,jiang2024hallucination}, and decoding refinements~\cite{zhuang2025vasparse,wang2024mitigating} have achieved partial mitigation of hallucinations, the problem remains substantial. 
This persistent issue highlights the need for reliable hallucination detection, especially in medical applications.

\begin{figure}[t]
\centering
\includegraphics[width=0.5\textwidth]{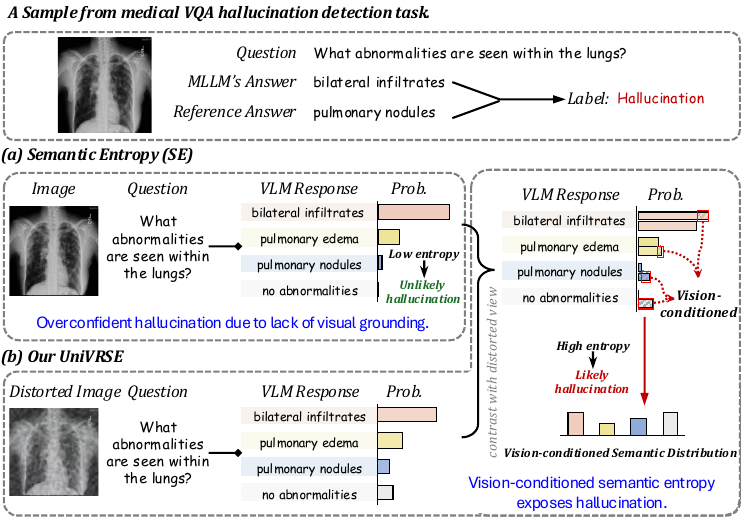}
\caption{
Motivation behind UniVRSE. UniVRSE contrasts the semantic predictive distributions derived from the original and visually distorted image–text pairs, enabling enhanced visual guidance and exposing hallucinations that remain undetected by SE.
}
\label{fig:intro_UniVRSE}
\end{figure}

Recent efforts on hallucination detection for LLMs and VLMs broadly fall into three categories:
(1) \textit{Supervised Detectors}, which train dedicated hallucination detection models but require costly annotated data and often fail to generalize to unseen scenarios~\cite{gunjal2024detecting,xiao2024detecting,chen2024detecting,hardy2024rextrust};
(2) \textit{External-Verification Methods}, which rely on additional sources of information, such as cross-examining outputs with other LLMs or VLMs~\cite{yu2024hallucidoctor,cohen2023lm}, double-checking with vision expert models~\cite{sahu2024pelican,yin2024woodpecker}, or verifying responses using external knowledge bases~\cite{chen2024complex,min2023factscore}; and
(3) \textit{Uncertainty-based Methods}, which detects hallucinations by estimating predictive uncertainty to assess hallucination likelihood~\cite{li2024reference,farquhar2024detecting,chen2024inside}.
Among them, uncertainty-based approaches offer a particularly appealing trade-off between scalability and model efficiency~\cite{liu2024mllmhallusurvey,huang2023llmhallusurvey,zou2023review}, requiring no auxiliary models, external knowledge bases, or task-specific fine-tuning.
A notable advancement is \textit{\textbf{Semantic Entropy (SE)}}~\cite{farquhar2024detecting}, originally proposed for LLMs, which estimates semantic-level predictive uncertainty by clustering diverse responses that convey the same meaning.
While SE achieves competitive results in text-only settings, its extension to VLMs has proved difficult~\cite{zhang2024vl}. 
Medical VLMs frequently exhibit \textit{\textbf{overconfident hallucinations}}, driven by strong linguistic priors and insufficient visual grounding, resulting in unreliable semantic uncertainty estimates. 
As shown in Fig.~\ref{fig:intro_UniVRSE}, responses can remain confidently incorrect even under perturbed visual content, revealing a modality bias that SE fails to address.

To address this limitation, we propose \textbf{UniVRSE}, a \textbf{Uni}fied \textbf{V}ision-conditioned \textbf{R}esponse \textbf{S}emantic \textbf{E}ntropy framework that explicitly strengthens visual guidance in semantic uncertainty estimation. 
UniVRSE is designed to be model-agnostic and applicable to both short-form VQA and long-form VRG.
For medical VQA, UniVRSE samples multiple responses for an image–question pair to derive a semantic predictive distribution, and then contrasts it against the distribution obtained from a visually distorted version of the same image. The entropy of this discrepancy distribution, termed \textit{vision-conditioned semantic entropy}, reflects the sensitivity of the response to visual information, providing a more robust indicator of hallucination.
For medical VRG, we further decompose the generated report into atomic claims, construct aligned image–question–claim triplets, and apply the same vision-conditioned estimation procedure at the claim level.
A higher vision-conditioned semantic entropy indicates a greater likelihood of hallucination.
A major challenge in benchmarking hallucination detection in medical VLMs lies in the \textit{\textbf{lack of reliable hallucination labels}}. Existing medical VQA and VRG datasets contain only reference responses, without annotations regarding factual correctness of model-generated outputs. Moreover, current hallucination evaluation metrics often rely on subjective or modality specific criteria, yielding limited reproducibility and poor cross-modality generalization. 
To overcome this, we introduce the \textbf{AL}ignment ratio of atomic \textbf{FA}cts (\textbf{ALFA}), a \textit{\textbf{fine-grained and objective evaluation metric}} that decomposes both generated and reference responses into atomic facts or claims and performs semantic alignment to quantify factual consistency. The resulting hallucination labels serve as robust supervision for benchmarking hallucination detection.
Extensive experiments have been conducted on four VQA datasets (MIMIC-Diff-VQA, Path-VQA, SLAKE, and RAD-VQA) and two VRG datasets (IU-Xray and CheXpertPlus) spanning diverse imaging modalities (CT, MRI, and pathology). 
Our results demonstrate that UniVRSE consistently outperforms recently competing detection methods and generalizes across three representative medical VLMs (\emph{i.e.}, MedGemma-4B-it~\cite{sellergren2025medgemma}, LLaVAMed-7B~\cite{li2024llavamed}, and HuaTuoGPT-Vision-7B~\cite{chen2024towards}).

The main contributions of this work are four-fold.
(1) We propose UniVRSE, a unified and model-agnostic framework for hallucination detection in medical VLMs that explicitly enhances visual guidance in semantic predictive uncertainty estimation.
(2) We address the modality bias overlooked by conventional Semantic Entropy by contrasting semantic distributions derived from original and distorted images, improving robustness to overconfident hallucinations.
(3) We introduce ALFA, a fine-grained and objective metric that evaluates factual consistency and enables automatic hallucination labeling across VQA and VRG tasks.
(4) Comprehensive experiments across diverse imaging modalities and VLM backbones demonstrate the superior performance and strong generalizability of UniVRSE in both short-form and long-form medical text generation tasks.

\section{Related Work}
\label{sec:relatedwork}
\subsection{Hallucination Detection in LLMs}

Hallucination detection in LLMs has been studied from several complementary perspectives. 
A first paradigm is \textit{supervised detector-based approaches}, where classification heads or LLMs are fine-tuned to distinguish factual responses from hallucinated ones~\cite{han2024semantic}. These methods can achieve strong performance given sufficient annotations but often display limited transferability to unseen domains due to dependence on task-specific training data. 
A second major direction centers on \textit{cross-model examination}, where outputs from multiple LLMs are compared to identify inconsistent predictions \cite{cohen2023lm,zhang2023sac}. Agreement among models is deemed reliable, while divergence may indicate hallucination. However, performance is inherently limited by the quality and diversity of the reference models.
A third line of research leverages \textit{external knowledge retrieval}, wherein generated responses are verified against structured external knowledge bases~\cite{varshney2023stitch,sun2025redeep} or web evidence~\cite{chen2024complex}. 
While external evidence provides explicit factual grounding, performance depends heavily on retrieval precision and domain coverage, which can be inconsistent across clinical contexts.
Finally, \textit{uncertainty or consistency estimation}, which seeks to quantify hallucination risk using the model’s own predictive behavior~\cite{li2024reference,kuhnsemantic,farquhar2024detecting,manakul2023selfcheckgpt,chen2024inside,agrawal2024language,varshney2023stitch}. 
Methods in this family estimate uncertainty at different granularities, including token-level~\cite{li2024reference,varshney2023stitch}, semantic-level (\emph{e.g.}, Semantic Entropy)~\cite{kuhnsemantic,farquhar2024detecting,manakul2023selfcheckgpt,agrawal2024language}, and embedding-level~\cite{chen2024inside}. 
These approaches are model-efficient, require no external supervision, and operate without auxiliary networks or retrieval modules, making them particularly suitable for scalable deployment.
Overall, uncertainty-based strategies have emerged as a particularly promising solution for reliable hallucination detection, offering a balance between scalability, model efficiency, and general applicability.

\subsection{Hallucination Detection in VLMs}
Hallucination detection in VLMs largely follows the paradigms established in text-only LLMs, including
supervised detector-based approaches~\cite{gunjal2024detecting,whitehead2024pre,xiao2024detecting}, 
cross-model examination~\cite{yu2024hallucidoctor}, 
self-consistency analysis~\cite{wu2024logical,zhang2024dhcp}, and predictive uncertainty estimation~\cite{zhang2024vl}.
In the medical domain, recent efforts have further adapted these strategies, including employing self-consistency analysis~\cite{sambara2024radflag}, fine-tuning medical MLLMs with annotated hallucination data~\cite{chen2024detecting}, or training a hallucination
classification head~\cite{hardy2024rextrust}. 
A research direction particularly relevant to VLMs emphasizes \textbf{visual grounding verification}, wherein expert vision models assess whether generated responses align with visual evidence~\cite{yin2024woodpecker,chen2024unified,kim2024esreal,park2025convis,sahu2024pelican,zhang2024meter}. While these approaches provide explicit visual reasoning signals, they introduce computational overhead and dependency on the reliability of external expert models. By contrast, \textit{uncertainty-based approaches} rely only on internal model confidence and thus avoid additional model dependencies.
To extend uncertainty estimation to multimodal settings, VL-Uncertainty~\cite{zhang2024vl} adapts \textit{Semantic Entropy} by applying transformations to the visual input when sampling responses. However, VLMs frequently exhibit \textit{\textbf{modality bias}}. Their predictions are dominated by linguistic priors learned from large-scale text corpora, resulting in overconfident hallucinations that persist even under visual perturbations.
Our prior work, VASE~\cite{liao2025VASE}, addressed this issue for short-form VQA but was limited to single-answer scenarios and lacked extensibility to long-form generation.
In this work, we introduce \textbf{UniVRSE}, a unified uncertainty-based framework that explicitly enhances visual guidance in semantic predictive entropy. UniVRSE generalizes across VQA and VRG settings and achieves strong performance across multiple datasets and medical VLM backbones. 
By mitigating overconfidence and exploiting vision-conditioned semantic divergence, UniVRSE enhances the reliability of uncertainty estimation for hallucination detection in medical VLMs.

\section{Hallucination Label Acquisition via ALFA}
\label{sec:alfa_details}
Reliable hallucination detection requires ground-truth labels indicating whether model-generated responses are factually aligned with reference answers. However, medical VQA and VRG datasets typically provide only multimodal inputs and reference responses, without annotations on factual correctness or hallucination presence for VLM-generated outputs.
To construct a benchmark suitable for hallucination detection, we first generate responses using a medical VLM and then determine hallucination labels by comparing each generated response with its corresponding reference answer via hallucination evaluation metrics. 
Thus, the acquisition of reliable hallucination labels becomes a crucial prerequisite for evaluating hallucination detection.
Below, we summarize existing hallucination evaluation metrics and highlight their limitations before introducing our proposed metric, ALFA.

\subsection{Existing Metrics for Evaluating Factual Correctness}
Traditional text similarity metrics, such as BLEU~\cite{papineni2002bleu}, ROUGE~\cite{lin2004rouge}, METEOR~\cite{banerjee2005meteor}, and BERT-Score~\cite{zhang2019bertscore}, quantify lexical overlap or embedding-level similarity but do not capture fine-grained factual correctness.
Extensions based on domain-specific overlap, including disease matching~\cite{chang2025medheval} or anatomical component alignment~\cite{zuo2025medhallbench}, offer improved clinical relevance yet still overlook essential attributes and inter-entity relations. As a result, they fail to detect subtle inconsistencies or spurious statements.

Recent studies have explored \textbf{LLM-based evaluators} to assess factual consistency. For instance, \textit{MedHallTune}~\cite{yan2025medhalltune} employs GPT-4o to compare the generated and reference responses, assigning a score from 1 to 10 across four aspects: (i) clinical precision, (ii) clinical relevance, (iii) detail level, and (iv) risk level. 
However, such grading remains subjective and exhibits high sensitivity to prompt design and evaluation model selection, leading to limited reproducibility.
\textit{MediHall Score}~\cite{chen2024detecting} develops a taxonomy of hallucinations with severity levels, including \textit{Catastrophic}, \textit{Critical}, \textit{Attribute}, \textit{Prompt-induced}, and \textit{Minor Hallucinations}, but its categorization criteria are ambiguous and lack quantitative grounding, making consistent annotation difficult. For example, the boundary between \textit{Catastrophic} and \textit{Critical} hallucinations is ill-defined, and descriptions such as `affect diagnostic accuracy to some extent' or `not seriously affect clinical diagnosis' provide no quantitative basis for consistent annotation.
Another notable effort, \textit{GREEN}~\cite{ostmeier2024green} is an LLM that is trained on large-scale radiological datasets and evaluates radiology reports across six error types, including:
(i) false report of a finding in the candidate,
(ii) missing a finding present in the reference,
(iii) misidentification of a finding's anatomic location/position,
(iv) misassessment of the severity of a finding,
(v) mentioning a comparison that isn't in the reference, and
(vi) omitting a comparison detailing a change from a prior study.
Nevertheless, its reliability declines markedly when applied to other modalities or heterogeneous clinical narratives (see Section~\ref{sec:alfa_cross_modal}). 

Overall, existing factuality metrics suffer from two fundamental drawbacks: \textit{\textbf{(i) ambiguous or subjective evaluation principles, and (ii) poor generalization across imaging modalities and clinical domains}}. These limitations motivate the need for an objective and scalable metric that captures factual alignment at a clinically meaningful level.

\subsection{ALFA Metric}

\begin{figure}[t]
\centering
\includegraphics[width=0.5\textwidth]{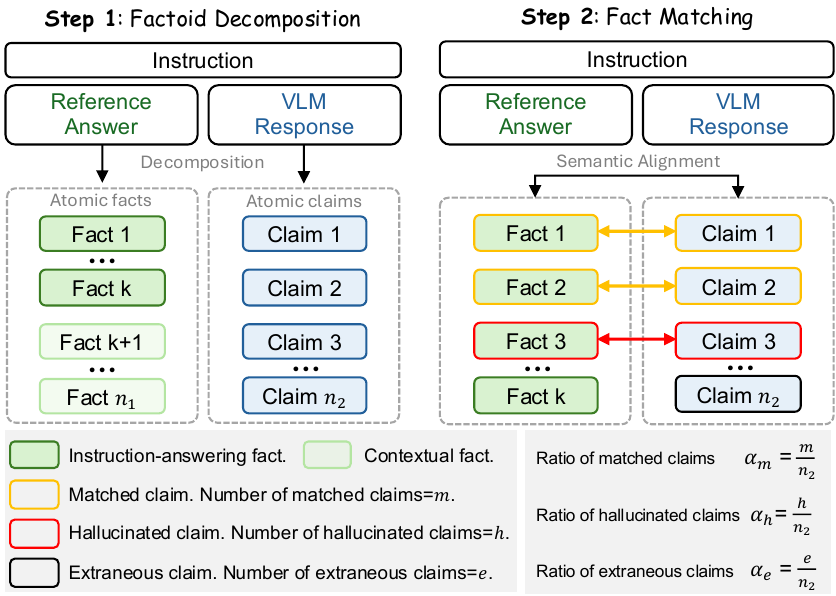}
\caption{
Overview of ALFA evaluation framework. ALFA provides fine-grained hallucination assessment by performing atomic fact–level alignment between the response generated by the medical VLM and the corresponding reference answer.
}
\label{fig:alfa_intro}
\end{figure}

To address these shortcomings, we propose \textit{ALFA}, a fine-grained and interpretable metric for evaluating factual consistency in open-ended medical text generation. As illustrated in Fig. ~\ref{fig:alfa_intro}, ALFA compares a low-temperature ($T$ = 0.1) sampled response from a medical VLM~\cite{farquhar2024detecting} with the reference answer for each sample. The evaluation process consists of two main components: factoid decomposition and fact matching.

\textbf{Factoid Decomposition.}
We first decompose both the reference answer and VLM response into atomic semantic units. For the reference answer, each unit represents a clinically verified and self-contained \textit{atomic facts}.
For the generated response, the units are referred to as \textit{atomic claims}, representing assertions whose validity must be assessed. Let the reference answer contain $n_1$ atomic facts and the generated response contain $n_2$ atomic claims. Reference facts are further divided into \textit{instruction-answering facts} and \textit{contextual facts}.
Contextual facts are derived from background information relevant to the instruction, whereas instruction-answering facts directly address the instruction and serve as the primary basis for evaluating the VLM response.

\textbf{Fact Matching.}
Each atomic claim is then aligned with reference facts and categorized as one of:
\begin{itemize}
    \item \textit{Matched claims}: semantically consistent with reference facts; 
    \item \textit{Hallucinated claims}: contradicted by or unsupported in the reference answer; 
    \item \textit{Extraneous claims}: introduce additional information not mentioned in the reference facts.
\end{itemize}
Let $m$, $h$, and $e$ denote the number of matched, hallucinated, and extraneous claims, respectively. 
ALFA computes the ratios of these three kinds of claims in claims of VLM response as follows:
\begin{equation}
    \boldsymbol{\alpha}_{\textbf{m}} = \frac{m}{n_2},\\ 
    \boldsymbol{\alpha}_{\textbf{h}} = \frac{h}{n_2}, \\
    \boldsymbol{\alpha}_{\textbf{e}} = \frac{e}{n_2},
\end{equation}
where $\boldsymbol{\alpha}_{\textbf{m}}$, $\boldsymbol{\alpha}_{\textbf{h}}$, and $\boldsymbol{\alpha}_{\textbf{e}}$ 
are all within the range $[0,1]$ and satisfy $\boldsymbol{\alpha}_{\textbf{m}} + \boldsymbol{\alpha}_{\textbf{h}} + \boldsymbol{\alpha}_{\textbf{e}} = 1$.

In this study, $\boldsymbol{\alpha}_{\textbf{h}}$ is used as the hallucination label in our evaluation. In practice, both factoid decomposition and fact matching are performed using DeepSeek-V3~\cite{liu2024deepseek}. 
By operating at the atomic-fact level, ALFA provides a structured and clinically interpretable evaluation of hallucinations, enabling reproducible annotation and cross-modality generalization. A detailed analysis is provided in Sections~\ref{sec:alfa_eval} and \ref{sec:alfa_cross_modal}.

\begin{figure*}[t]
\includegraphics[width=1.0\textwidth]{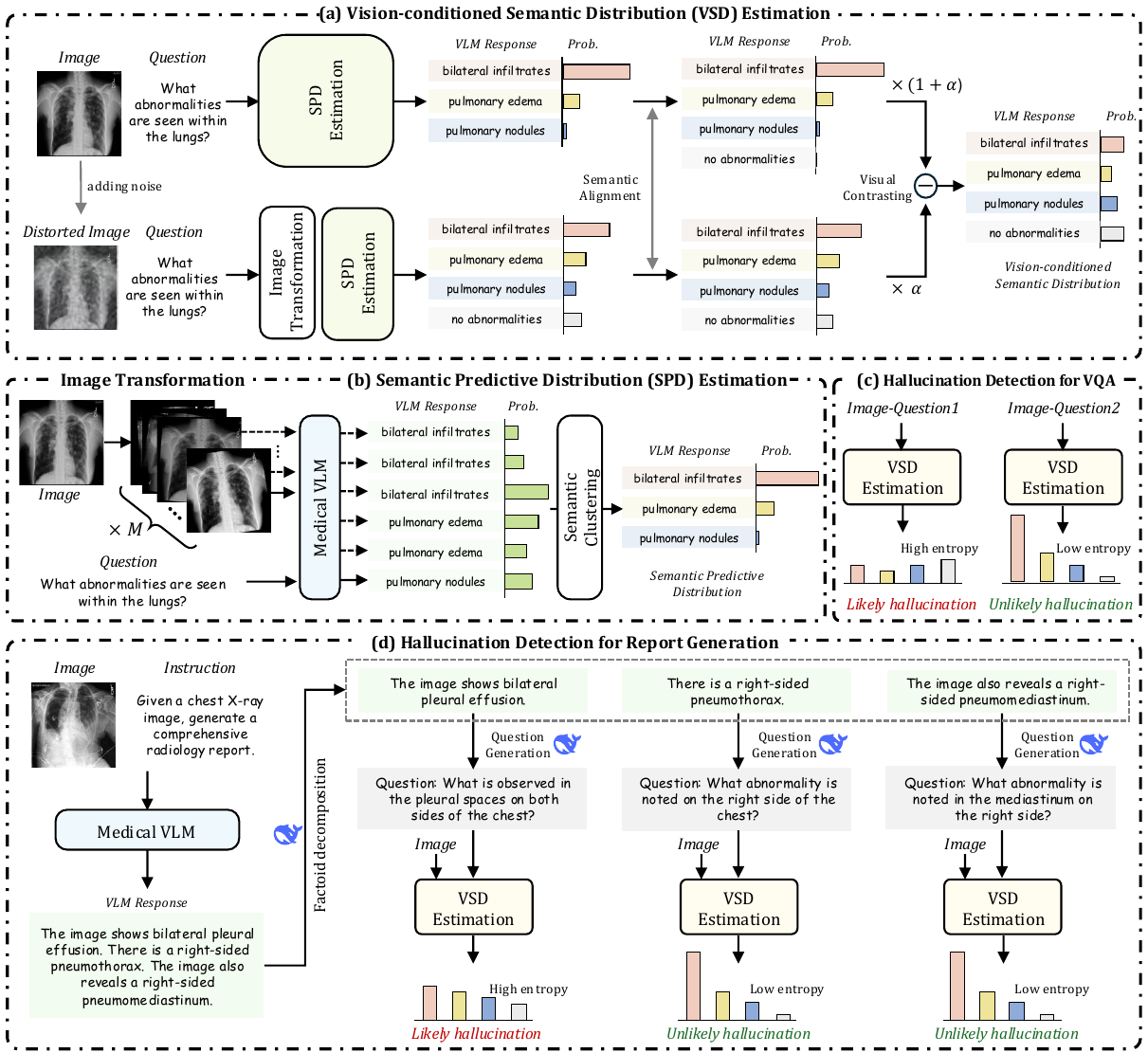}
\caption{
Architecture of the proposed UniVRSE framework for hallucination detection in medical VQA and VRG tasks. `Alignment' denotes semantic equivalence class matching, and `Prob.' represents the estimated probability of each semantic class.
}
\label{fig:overview}
\end{figure*}

\section{Methodology}
\label{sec:method_UniVRSE}
\subsection{Problem Formulation and Framework Overview}

We present the UniVRSE framework, starting with its application to VQA and subsequently extending it to long-form VRG. 
In the VQA setting, let the multimodal input be a pair $(\textbf{\textit{x}}_v, \textbf{\textit{x}}_q)$, where $\textbf{\textit{x}}_v$ is the medical image and $\textbf{\textit{x}}_q$ is the textual query. A medical VLM $f$ generates a response $\textbf{\textit{r}}$. 
Our objective is to determine if $\textbf{\textit{r}}$ contains statements that contradict the visual evidence, which we define as hallucinated content.

UniVRSE addresses this by estimating the \textbf{Vision-conditioned Semantic Entropy (VSE)} of the response (see Fig.~\ref{fig:overview}). Given $(\textbf{\textit{x}}_v, \textbf{\textit{x}}_q)$, we first derive a Semantic Predictive Distribution (SPD) from the original image. This distribution is then contrasted with the SPD obtained using a visually distorted version of the same image. The entropy of this resultant discrepancy distribution serves as the VSE score. A response $\textbf{\textit{r}}$ is classified as a hallucination if its VSE exceeds an empirically determined threshold $\tau$. The UniVRSE procedure is formalized in three principal stages:
(1) Estimation of SPD.  
(2) Derivation of the Vision-conditioned Semantic Distribution (VSD). 
(3) Computation of the VSE from the VSD to quantify hallucination likelihood.
We delve into the details of each step.

\subsection{SPD Estimation}
\label{sec:SPD}

To obtain a reliable estimation of the semantic predictive distribution, we require diverse model outputs. We achieve this by applying a set of mild transformations to the input image $\mathbf{x}_v$, including random cropping, rotation, translation, contrast adjustment, and brightness adjustment. These perturbations induce variation in visual presentation while rigorously preserving clinical semantics (see Section~\ref{sec:ana_img_trans} for ablations). The SPD estimation follows a rigorous two-step procedure: Generation and Clustering.

\subsubsection{Generation Step} 
Using the transformed image $\mathbf{x}_v$ as inputs, we perform $M$ inference runs. The medical VLM $f$ generates $M$ output sequences $\{\textbf{\textit{s}}^{(i)}\}_{i=1}^{M}$ using high-temperature sampling to promote diverse responses.
The token-level probability for each generated sequence is recorded as:
\begin{equation}
    P(\textbf{\textit{s}}^{(i)}|\textbf{\textit{x}}_v, \textbf{\textit{x}}_q)
    = \prod_j P(s_j^{(i)}|\textbf{\textit{s}}_{<j}^{(i)}, \textbf{\textit{x}}_v, \textbf{\textit{x}}_q),
\end{equation}
where $s_j^{(i)}$ is the $j$-th token of $\textbf{s}^{(i)}$ and $\textbf{s}^{(i)}_{<j}$ represents its preceding tokens.

\subsubsection{Clustering Step} 
To approximate the semantic predictive distribution, the $M$ generated sequences $\{\textbf{\textit{s}}^{(i)}\}_{i=1}^{M}$ are clustered into \textbf{semantic equivalence classes} by evaluating bidirectional entailment relations~\cite{farquhar2024detecting}. We use a powerful Natural Language Inference model, DeBERTa-Large-MNLI~\cite{he2020deberta}, for this evaluation. Specifically, two sequences are considered semantically equivalent if they mutually entail each other. This sequential clustering procedure yields ${N_1}$ semantic equivalence classes $\{c_i\}_{i=1}^{N_1}$. The probability of each class $c_i$ is computed by aggregating the probabilities of its constituent sequences:
\begin{equation}
    P(c_i|\textbf{\textit{x}}_v, \textbf{\textit{x}}_q)
    = {\textstyle \sum_{\textbf{\textit{s}} \in c_i}} P(\textbf{\textit{s}}|\textbf{\textit{x}}_v, \textbf{\textit{x}}_q).
\end{equation}
This yields the initial SPD: $[P(c_i|\textbf{\textit{x}}_v, \textbf{\textit{x}}_q)]_{i=1}^{N_1}$.

\subsection{VSD Estimation}
\label{sec:CSPD}

To amplify the influence of visual evidence and suppress language-driven overconfidence, we introduce a contrasting mechanism. We generate a visually distorted image, $\mathbf{x}'_v$, by applying noise injections that degrade fine-grained details. We estimate a second SPD, $[P(c_i|\textbf{\textit{x}}_v', \textbf{\textit{x}}_q)]_{i=1}^{N_2}$, using this distorted image $\mathbf{x}'_v$ and the same query $\mathbf{x}_q$, 
where $N_2$ is the number of semantic equivalence classes derived from the distorted input.

Since the equivalence class sets from the two distributions, $\{c_i\}_{i=1}^{N_1}$ and $\{c_i\}_{i=1}^{N_2}$, may differ, we employ the same NLI model to align them into a unified semantic space containing $N$ classes, denoted by the set $\{c_i\}_{i=1}^{N}$. The aligned distributions are denoted as $[P(c_i|\textbf{\textit{x}}_v, \textbf{\textit{x}}_q)]_{i=1}^{N}$ and $[P(c_i|\textbf{\textit{x}}_v', \textbf{\textit{x}}_q)]_{i=1}^{N}$.
The VSD, $P_{dis}$, is then defined by contrasting these two probability vectors:
\begin{equation}
    P_{dis}(\textbf{\textit{c}}|\textbf{\textit{x}}_v,\textbf{\textit{x}}_q) = \sigma((1+\lambda)P(\textbf{\textit{c}}|\textbf{\textit{x}}_v,\textbf{\textit{x}}_q)- 
    \lambda P(\textbf{\textit{c}}|\textbf{\textit{x}}_v',\textbf{\textit{x}}_q)),
\end{equation}
where $\sigma(\cdot)$ denotes softmax normalization, 
and $\lambda \ge 0$ controls the strength of the visual amplification. 
This contrasting mechanism effectively suppresses predictions where $P(\textbf{\textit{c}}|\textbf{\textit{x}}_v,\textbf{\textit{x}}_q) \approx P(\textbf{\textit{c}}|\textbf{\textit{x}}_v',\textbf{\textit{x}}_q)$. Such similarity indicates that the VLM is relying heavily on linguistic priors, as the visual degradation ($\mathbf{x}'_v$) had minimal impact on the outcome. By contrast, claims that are highly visually grounded will show a large drop in probability when the image is distorted, thus achieving visual amplification in the VSD.

\subsection{Hallucination Detection for VQA}

The predictive uncertainty for the input pair is quantified by VSE, which is the entropy of the VSD:
\begin{equation}
\text{VSE}(\textbf{\textit{x}}_v,\textbf{\textit{x}}_q)
    = -{\textstyle \sum_{i=1}^{N}} P_{dis}(c_i|\textbf{\textit{x}}_v,\textbf{\textit{x}}_q)
    \log P_{dis}(c_i|\textbf{\textit{x}}_v,\textbf{\textit{x}}_q).
\end{equation}
A higher VSE score indicates greater uncertainty in the model's prediction, correlating with an increased likelihood of hallucinated content. An answer $\textbf{\textit{r}}$ is labeled as hallucinated if its VSE exceeds a threshold $\tau \in \mathcal{R}^+$, which is empirically determined on a held-out validation set~\cite{peng2025enhancing}.

It is important to note that VSE quantifies the model's predictive uncertainty for the input pair $(\textbf{\textit{x}}_v,\textbf{\textit{x}}_q)$, rather than directly assessing the factual correctness of the specific response $\textbf{\textit{r}}$. High uncertainty indicates low confidence in the prediction set, which aligns with the core principle of uncertainty-based detection frameworks, as validated by prior work and our experimental results.

\subsection{Hallucination Detection for VRG}
\label{sec:UniVRSE_vrg}
To address the challenge of hallucination in long-form, open-ended clinical reports, UniVRSE is extended to operate at the fine-grained \textit{factual claim} level.
Given a medical image $\textbf{\textit{x}}_v$ and an instruction prompt $\textbf{\textit{x}}_I$, the medical VLM generates a report $\textbf{\textit{r}}$ describing anatomical observations, pathological findings, and diagnostic impressions. 
Since a single report sentence often contains multiple factual assertions, we must decompose $\textbf{\textit{r}}$ into minimal semantic units, or \textit{atomic claims}, each expressing a distinct, self-contained clinical statement that can be independently verified.

The detection process for VRG utilizes two post-processing stages based on powerful LLMs:
(1) \textit{Factoid Decomposition:} The generated report $\textbf{\textit{r}}$ is segmented into a set of atomic claims $\{f_j\}_{j=1}^{J}$, where $J$ is the total number of claims; and 
(2) \textit{Question Generation:} For each atomic claim $f_j$, a verification query $\textbf{\textit{x}}_q^{(j)}$ is automatically formulated such that $f_j$ serves as the expected answer. This effectively converts each claim into an independent VQA-style verification instance. 
This decomposition transforms the report-level hallucination detection task into $J$ independent VQA subtasks, enabling fine-grained evidence verification. We utilize a powerful external LLM, such as \textit{DeepSeek-V3}~\cite{liu2024deepseek} or \textit{GPT-4o}~\cite{hurst2024gpt} , for these stages due to their advanced factual reasoning capability. In this work, we adopt \textit{DeepSeek-V3}.

As illustrated in Fig.~\ref{fig:overview}, each resulting image–question pair $(\textbf{\textit{x}}_v, \textbf{\textit{x}}_q^{(j)})$ is processed by the core VSD module, and its VSE is computed following Section~\ref{sec:CSPD}. 
The resulting VSE score measures the \textit{degree of visual support} for the corresponding atomic claim $f_j$. A high VSE indicates that the claim is weakly grounded and potentially hallucinated, while low entropy suggests strong alignment with the visual evidence. This fine-grained verification strategy is demonstrated in the subsequent experiments to be critical for achieving clinically reliable report generation.

\section{Experiments}
\label{sec:exp_all}
\subsection{Datasets}

We evaluated UniVRSE on four open-ended VQA datasets and two VRG datasets that together span diverse imaging modalities and clinical reasoning tasks. The VQA benchmarks include VQA-RAD~\cite{lau2018dataset}, SLAKE~\cite{liu2021slake}, Path-VQA~\cite{he2020pathvqa}, and MIMIC-Diff-VQA~\cite{hu2023expert}, while VRG evaluation is conducted on CheXpertPlus~\cite{chambon2024chexpertplus} and IU-Xray~\cite{demner2015preparing}.

\textbf{CheXpertPlus}~\cite{chambon2024chexpertplus} comprises 223,228 chest X-ray–report pairs from 64,725 patients. Since the official test set is unavailable, we followed prior work and evaluated hallucination detection using the 234 samples in the validation set. 
\textbf{IU-Xray}~\cite{demner2015preparing} contains 2,955 chest X-ray–report pairs organized into 2,069/296/590 train/val/test samples. We employed the official test split for evaluation.

\textbf{VQA-RAD}~\cite{lau2018dataset} includes 2,244 question–answer pairs paired with 314 radiology images, covering both open-ended and binary (`yes/no') questions. We evaluated hallucination detection only on the 200 open-ended questions in the test split.
\textbf{SLAKE}~\cite{liu2021slake} is a multimodal and bilingual medical VQA benchmark spanning CT, MRI, and X-ray images and multiple anatomical regions such as the head, neck, and chest. 
It includes both open-ended and closed-ended questions, with 9,835 training, 2,099 validation, and 2,094 test samples. 
For evaluation, we selected only English open-ended questions from the test set, yielding 645 samples.
\textbf{Path-VQA}~\cite{he2020pathvqa} consists of question–answer pairs collected from pathology images, covering both open-ended and binary question types. 
It contains 19,654 training samples, 6,259 validation samples, and 6,719 test samples. 
We evaluated UniVRSE on the open-ended subset of the test split, which includes 3,357 samples.
\textbf{MIMIC-Diff-VQA}~\cite{hu2023expert} is a large-scale chest X-ray VQA dataset containing 700,703 question–answer pairs across seven categories: abnormality, difference, level, location, presence, type, and view. 
To prevent data leakage, as several medical VLMs use MIMIC-CXR for pretraining, we followed the MIMIC-CXR training split~\cite{johnson2019mimic} instead of the official one, yielding 13,121 test samples. 
For open-ended hallucination detection, we excluded `view' questions with fixed answers (\textit{e.g.}, `AP view' or `PA view')
and all binary `yes/no' samples, resulting in a final open-ended test set of 7,908 samples.

\begin{table*}[t]
\caption{
Hallucination detection performance (AUC(\%) and AUA(\%)) of UniVRSE and seven competing methods on open-ended VQA benchmarks. Best results are shown in \textbf{bold}, and the second-best results are \underline{underlined}.
}
\label{tab:comparison_vqa}
\centering
\setlength\tabcolsep{15pt}
\begin{tabular}{c|c|c|c|c|c|c|c|c}
\hline \hline
\multirow{2}{*}{Method} & \multicolumn{2}{c|}{RAD-VQA}  & \multicolumn{2}{c|}{SLAKE} & \multicolumn{2}{c|}{Path-VQA}    & \multicolumn{2}{c}{MIMIC-Diff-VQA} \\ \cline{2-9} 
                & AUC $\uparrow$ & AUA $\downarrow$ & AUC $\uparrow$ & AUA $\downarrow$ & AUC $\uparrow$ & AUA $\downarrow$ & AUC $\uparrow$ & AUA $\downarrow$ \\ \hline 
\multicolumn{9}{c}{MedGemma-4B-it~\cite{sellergren2025medgemma}}                                  \\ \hline
AvgProb         & 40.87 & 40.98 & 43.29 & 24.04 & 48.21 & 40.23 & 49.04 & \underline{37.31} \\ 
AvgEnt          & 59.21 & 26.94 & 56.58 & 25.86 & 52.40 & 39.65 & 50.89 & 40.35 \\
MaxProb         & 41.06 & 38.16 & 42.35 & 26.36 & 48.44 & 39.87 & 49.27 & \textbf{37.16} \\ 
MaxEnt          & 58.83 & 27.05 & 57.30 & 22.76 & 51.66 & 40.59 & 50.39 & 40.72 \\ 
Cross-Checking  & 64.64 & 24.75 & 65.50 & 18.17 & 54.01 & 36.86 & 50.44 & 41.89 \\ 
RadFlag         & 70.15 & \underline{23.00} & 67.43 & \textbf{16.85} & 56.48 & 35.11 & \textbf{52.51} & 40.60 \\ 
SE              & \underline{71.87} & 24.79 & \underline{67.93} & \underline{17.44} & \underline{57.19} & \underline{34.88} & \underline{52.42} & 40.76 \\ 
UniVRSE            & \textbf{76.25} & \textbf{18.22} & \textbf{69.56} & 17.67 & \textbf{59.17} & \textbf{32.98} & 52.04 & 40.42 \\ \hline 
\multicolumn{9}{c}{LlavaMed-7B~\cite{li2024llavamed}}                                  \\ \hline
AvgProb         & 43.57 & 55.94 & 49.97 & 52.42 & 45.48 & 64.45 & 41.10 & 70.67 \\ 
AvgEnt          & 54.76 & 50.08 & 50.78 & 51.07 & 55.59 & 55.94 & 48.68 & 63.78 \\ 
MaxProb         & 41.71 & 55.18 & 49.75 & 51.91 & 48.94 & 59.97 & 40.71 & 69.76 \\ 
MaxEnt          & 56.02 & 51.08 & 52.18 & 53.41 & 53.01 & 60.02 & 49.20 & 64.51 \\ 
Cross-Checking  & 59.32 & 50.00 & 54.59 & 49.57 & 61.89 & 52.16 & 54.82 & 64.41 \\ 
RadFlag         & 69.46 & 40.72 & 61.00 & 45.64 & 61.56 & 52.77 & 56.10 & 62.42 \\ 
SE              & \underline{72.90} & \underline{40.54} & \underline{66.75} & \underline{43.79} & \underline{63.73} & \underline{50.60} & \underline{56.31} & \underline{62.34} \\ 
UniVRSE            & \textbf{74.31} & \textbf{39.02} & \textbf{68.16} & \textbf{42.07} & \textbf{66.18} & \textbf{48.72} & \textbf{58.13} & \textbf{61.05} \\ \hline
\multicolumn{9}{c}{HuatuoGPT-Vision-7B~\cite{chen2024towards}}                            \\ \hline
AvgProb         & 35.28 & 48.00 & 49.58 & 42.66 & 45.99 & 62.08 & 46.89 & 65.04 \\ 
AvgEnt          & 66.73 & 29.91 & 47.64 & 44.96 & 54.26 & 56.28 & 54.90 & 66.66 \\ 
MaxProb         & 34.20 & 47.13 & 47.35 & 44.28 & 46.14 & 61.45 & 46.61 & 64.17 \\ 
MaxEnt          & 69.01 & 29.28 & 52.07 & 43.92 & 55.42 & 56.47 & 56.05 & 66.32 \\ 
Cross-Checking  & 66.44 & 33.70 & 63.64 & 33.59 & 56.98 & 53.71 & \underline{58.54} & 60.94 \\ 
RadFlag         & 78.44 & 23.19 & 69.28 & \underline{30.29} & 57.81 & 54.04 & 58.11 & \underline{60.23} \\ 
SE              & \underline{79.02} & \underline{23.52} & \underline{69.60} & 30.42 & \underline{60.78} & \underline{52.96} & 58.34 & 60.51 \\ 
UniVRSE            & \textbf{81.17} & \textbf{22.05} & \textbf{70.89} & \textbf{29.06} & \textbf{62.89} & \textbf{51.18} & \textbf{60.71} & \textbf{59.19} \\ \hline \hline
\end{tabular}
\end{table*}

\subsection{Evaluation Metrics}
For each test sample, a response is first generated by the target medical VLM, and hallucination labels are assigned using ALFA. To evaluate hallucination detection methods, we adopt two metrics following~\cite{farquhar2024detecting,liao2025VASE}.
\textbf{AUC} measures the probability that a randomly chosen correct answer receives higher confidence (lower uncertainty) than a randomly chosen hallucinated answer.
Higher AUC indicate better hallucination detection performance.
\textbf{Area Under ALFA Curve (AUA)} summarizes the average hallucination degree $\boldsymbol{\alpha}_{\textbf{h}}$ among the top-X\% confident predictions. Specifically, we computed the mean hallucination score $\boldsymbol{\alpha}_{\mathbf{h}}$ for $X \in {1\%, \dots,100\%}$, and integrated the resulting curve. 
Lower AUA indicates stronger hallucination suppression. 
Together, AUC and AUA provide complementary views of hallucination detection methods' ranking ability.

\subsection{Implementation Details}
\label{sec:implementation}

We set the sampling number $\textit{M}=10$ and used temperature 1.0 to encourage response diversity. The visual amplification coefficient is fixed at $\lambda=1.0$.
The weak transformations used include:
random cropping (selecting 90\% to 100\% of the original area),
random rotation (choosing an angle between -10° and 10°),
random translation (displacing up to 10\% of the image's width or height), 
and brightness/contrast adjustments (uniformly selected between 0.8 and 1.2).
Distortion was introduced via Gaussian noise (mean=0, std=0.07) and Poisson noise (scaling factor=70).
All experiments were conducted on two NVIDIA RTX 3090 GPUs.

\subsection{Comparison With Existing Methods}

The proposed UniVRSE is compared with several recent hallucination detection methods, including:
(1) AvgProb and MaxProb~\cite{li2024reference} quantify model confidence by computing the mean and peak token probabilities within the generated response. Lower confidence values suggest a greater risk of hallucination;
(2) AvgEnt and MaxEnt~\cite{li2024reference} quantify uncertainty by calculating the mean and maximum entropy of the token probability distribution, with higher entropy suggesting a greater likelihood of hallucinations.
(3) SE~\cite{farquhar2024detecting} estimates semantic predictive uncertainty by sampling multiple responses, forming a semantic distribution over answers via semantic clustering, and computing the resulting entropy;
(4) RadFlag~\cite{sambara2024radflag} samples multiple responses and computes the proportion that agree with the original answer, where a lower agreement rate indicates a higher likelihood of hallucinations; and
(5) Cross-Checking~\cite{yu2024hallucidoctor} produces additional responses from other VLMs for the same input and computes the average consistency between these alternative outputs and the original response.

\textbf{Results on VQA.}
Table~\ref{tab:comparison_vqa} reports the performance of various hallucination detection methods on RAD-VQA, SLAKE, Path-VQA, and MIMIC-Diff-VQA datasets using three medical VLM backbones: MedGemma-4B-it~\cite{sellergren2025medgemma}, LLaVAMed-7B~\cite{li2024llavamed}, and HuaTuoGPT-Vision-7B~\cite{chen2024towards}. 
Token-level confidence and entropy methods (\textit{AvgProb}, \textit{MaxProb}, \textit{AvgEnt}, and \textit{MaxEnt}) provide limited discrimination, suggesting that local token statistics are insufficient to capture multimodal hallucinations.
\textit{Cross-Checking} performs better than token-level uncertainty methods, but its effectiveness heavily depends on the capability of the auxiliary VLMs used for verification. When the checking model is weaker than the generation model, its detection performance degrades substantially. 
RadFlag, as a self-consistency-based method, performs comparably to but generally below the second-best baseline, Semantic Entropy. 
Across datasets and backbones, UniVRSE consistently achieves the highest AUC and lowest AUA.
In most cases, UniVRSE yields around a 2\% improvement in AUC over the second-best method, SE.
Specifically, on the RAD-VQA dataset using MedGemma-4B-it, UniVRSE surpasses SE by 4.38\% in AUC and 6.57\% in AUA.
Similarly, on the Path-VQA dataset with LlavaMed-7B, UniVRSE achieves gains of 2.45\% and 1.88\% in AUC and AUA, respectively.
This indicates that amplifying the contribution of visual evidence enhances the reliability of uncertainty estimation in identifying hallucinated content.

\textbf{Results on VRG.}
Table~\ref{tab:comparison_vrg} reports the atomic claim-level hallucination detection performance on two medical report generation datasets, CheXpertPlus and IU-Xray, using MedGemma-4B-it, LLaVAMed-7B, and HuaTuoGPT-Vision-7B. 
Compared with existing baselines, UniVRSE consistently achieves the best or near-best performance across all datasets and backbones. 
For example, on the IU-Xray dataset with MedGemma-4B-it, UniVRSE improves SE by 5.28\% in AUC and 2.69\% in AUA.
On the CheXpertPlus dataset with LlavaMed-7B, UniVRSE achieves improvements of 2.35\% and 1.11\% in AUC and AUA, respectively.
This demonstrates that amplifying visual evidence remains effective even under report-level complexity. 
Taken together, these results show that reinforcing visual evidence provides a stable basis for factuality estimation across a wide range of medical VLMs.

\begin{table}[t]
\caption{
Hallucination detection performance (AUC(\%) and AUA(\%)) of UniVRSE and competing methods on visual report generation benchmarks. Best results are shown in \textbf{bold}, and the second-best results are \underline{underlined}.
}
\label{tab:comparison_vrg}
\centering
\setlength\tabcolsep{10pt}
\begin{tabular}{c|c|c|c|c}
\hline \hline
\multirow{2}{*}{Method} & \multicolumn{2}{c|}{CheXpertPlus}  & \multicolumn{2}{c}{IU-Xray}  \\ \cline{2-5} 
                & AUC $\uparrow$ & AUA $\downarrow$ & AUC $\uparrow$ & AUA $\downarrow$ \\ \hline 
\multicolumn{5}{c}{MedGemma-4B-it~\cite{sellergren2025medgemma}}              \\ \hline
AvgProb         & 45.73 & 53.40 & 45.87 & 19.39 \\ 
AvgEnt          & 54.16 & 47.10 & 53.97 & 17.11 \\ 
MaxProb         & 45.30 & 55.10 & 44.96 & 21.16 \\ 
MaxEnt          & 54.49 & 46.86 & 54.80 & 17.22 \\ 
Cross-Checking  & 56.16 & 46.54 & 56.47 & 17.77 \\ 
RadFlag         & 56.53 & \textbf{44.53} & 56.12 & 16.09 \\ 
SE              & \underline{56.79} & \underline{44.61} & \underline{57.78} & \underline{15.07} \\ 
UniVRSE            & \textbf{58.96} & 44.71 & \textbf{63.06} & \textbf{12.38} \\ \hline 
\multicolumn{5}{c}{LlavaMed-7B~\cite{li2024llavamed}}                 \\ \hline
AvgProb         & 47.29 & 95.38 & 50.06 & 94.52 \\ 
AvgEnt          & 55.32 & 94.27 & 45.33 & 94.44 \\
MaxProb         & 48.31 & 95.41 & 43.81 & 95.30 \\ 
MaxEnt          & 56.26 & 94.13 & 58.11 & 93.83 \\ 
Cross-Checking  & 53.91 & 94.07 & 72.40 & 90.32 \\ 
RadFlag         & 55.70 & 94.13 & 69.65 & 90.44 \\ 
SE              & \underline{60.19} & \underline{93.97} & \underline{75.70} & \underline{90.14} \\ 
UniVRSE            & \textbf{62.52} & \textbf{93.08} & \textbf{78.06} & \textbf{89.03} \\ \hline
\multicolumn{5}{c}{HuatuoGPT-Vision-7B~\cite{chen2024towards}}         \\ \hline
AvgProb         & 43.02 & 59.69 & 45.93 & 36.39 \\ 
AvgEnt          & 55.08 & 50.48 & 52.69 & 32.21 \\ 
MaxProb         & 41.22 & 61.59 & 41.14 & 42.16 \\ 
MaxEnt          & 58.77 & 48.85 & 60.43 & 29.04 \\ 
Cross-Checking  & 55.68 & 48.93 & 56.77 & 28.93 \\ 
RadFlag         & 58.04 & 48.87 & 73.69 & 20.35 \\ 
SE              & \underline{60.18} & \underline{48.71} & \underline{80.39} & \underline{18.14} \\ 
UniVRSE            & \textbf{61.74} & \textbf{48.00} & \textbf{81.69} & \textbf{17.07} \\ \hline \hline
\end{tabular}
\end{table}

\begin{table}[t]
\caption{
Ablation study on the RAD-VQA dataset using MedGemma-4B-it. Hallucination detection performance (AUC(\%) and AUA(\%)) of UniVRSE and its variants. Best results are shown in \textbf{bold}.
}
\label{tab:ablation}
\centering
\setlength\tabcolsep{7pt}
\begin{tabular}{cc|c|c}
\hline \hline
\multicolumn{2}{c|}{UniVRSE}                                      & \multirow{2}{*}{AUC $\uparrow$} & \multirow{2}{*}{AUA $\downarrow$} \\ \cline{1-2}
\multicolumn{1}{c|}{Image Transformation} & Visual Contrasting &                      &                      \\ \hline 
\multicolumn{1}{c|}{$\times$}             & $\times$           &  71.87               &  24.79               \\ 
\multicolumn{1}{c|}{$\surd$}              & $\times$           &  71.94               &  24.16               \\ 
\multicolumn{1}{c|}{$\times$}             & $\surd$            &  74.37               &  19.70               \\ 
\multicolumn{1}{c|}{$\surd$}              & $\surd$            &  \textbf{76.25}               &  \textbf{18.22}               \\ \hline \hline
\end{tabular}
\end{table}

\subsection{Ablation Studies}

We further examined how each component of UniVRSE contributes by running ablations on \textit{visual contrasting} and \textit{image transformation}. 
In the \textit{Baseline} setting, semantic entropy is computed using only the original image–text pair for hallucination detection.
The \textit{image transformation (Baseline + IT)} extends the baseline by introducing visual diversity into uncertainty estimation. It estimates semantic entropy from randomly augmented images while keeping the textual input fixed.
The \textit{visual contrasting (Baseline + VC)} enhances the baseline by explicitly comparing the semantic predictive distributions of the original and distorted images, using the entropy of the resulting discrepancy distribution for hallucination detection. 
\textit{UniVRSE (Baseline + IT + VC)} integrates both visual contrasting and image transformation, combining their complementary effects for more reliable hallucination detection.
As shown in Table~\ref{tab:ablation}, both components significantly contribute to the overall performance. The Image Transformation variant yields a modest improvement in AUA by effectively introducing visual diversity. However, Visual Contrasting introduces a more substantial gain: a 2.5\% increase in AUC and 5.09\% increase in AUA. This result strongly highlights the benefit of explicitly contrasting model responses generated from original and visually-perturbed inputs. 
When combined, UniVRSE achieves the best overall performance, surpassing the baseline by 4.38\% in AUC and 6.57\% in AUA. This outcome confirms that the two modules work complementarily to substantially enhance the reliability of uncertainty estimation.

\begin{figure}[t]
\centering
\includegraphics[width=0.45\textwidth]{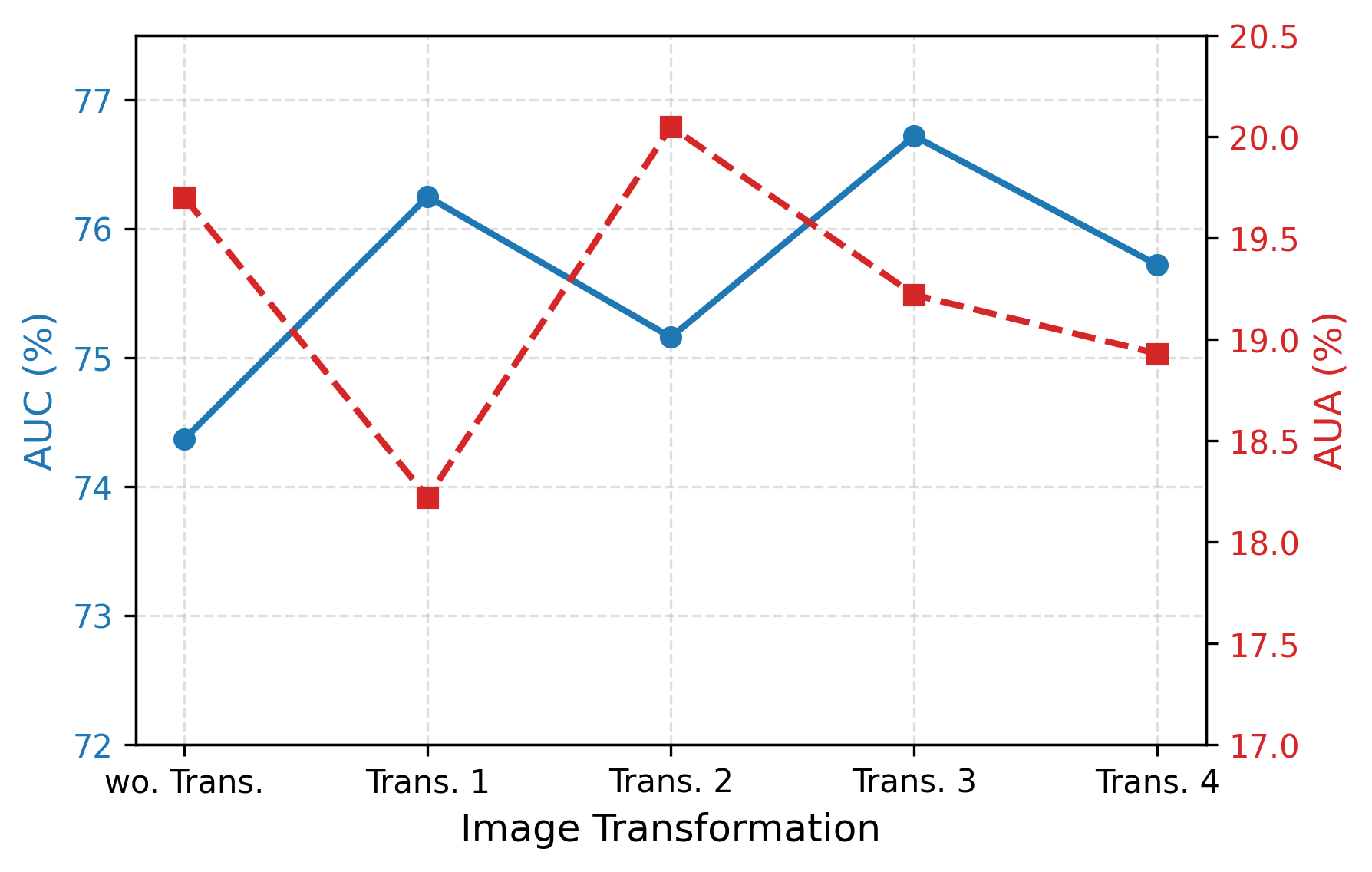}
\caption{
AUC (\%) and AUA (\%) of UniVRSE under different image transformation intensities on the RAD-VQA dataset (MedGemma-4B-it). The solid line and dashed line correspond to AUC and AUA, respectively.
}
\label{fig:image_trans_UniVRSE}
\end{figure}

\section{Discussion}

\subsection{Effect of Image Transformation Intensity}
\label{sec:ana_img_trans}
We investigated how visual perturbation intensity influences UniVRSE on RAD-VQA with MedGemma-4B-it. In this experiment, image transformations were applied exclusively to the distorted image, leaving the original image unaltered to serve as a reliable semantic anchor. Such transformations preserve the clinical semantics of the input while introducing controlled visual perturbations for contrastive estimation. 
Four transformation levels (Trans.~1–4) were tested with gradually increased degrees of cropping, rotation, color jittering, and translation (see Fig.~\ref{fig:image_trans_UniVRSE}). 
Introducing transformation yields measurable performance gains over the setting without augmentation (AUC: 74.37\%→76.25\%, AUA: 19.70\%→18.22\%), indicating that controlled visual randomness strengthens uncertainty estimation by enhancing contrastive signal diversity. Across different intensities, performance variations remain limited, suggesting that UniVRSE is insensitive to hyperparameter selection within a moderate range. Thus, Trans.~1 was adopted as the default configuration (See Section~\ref{sec:implementation}).

\begin{figure}[t]
\centering
\includegraphics[width=0.45\textwidth]{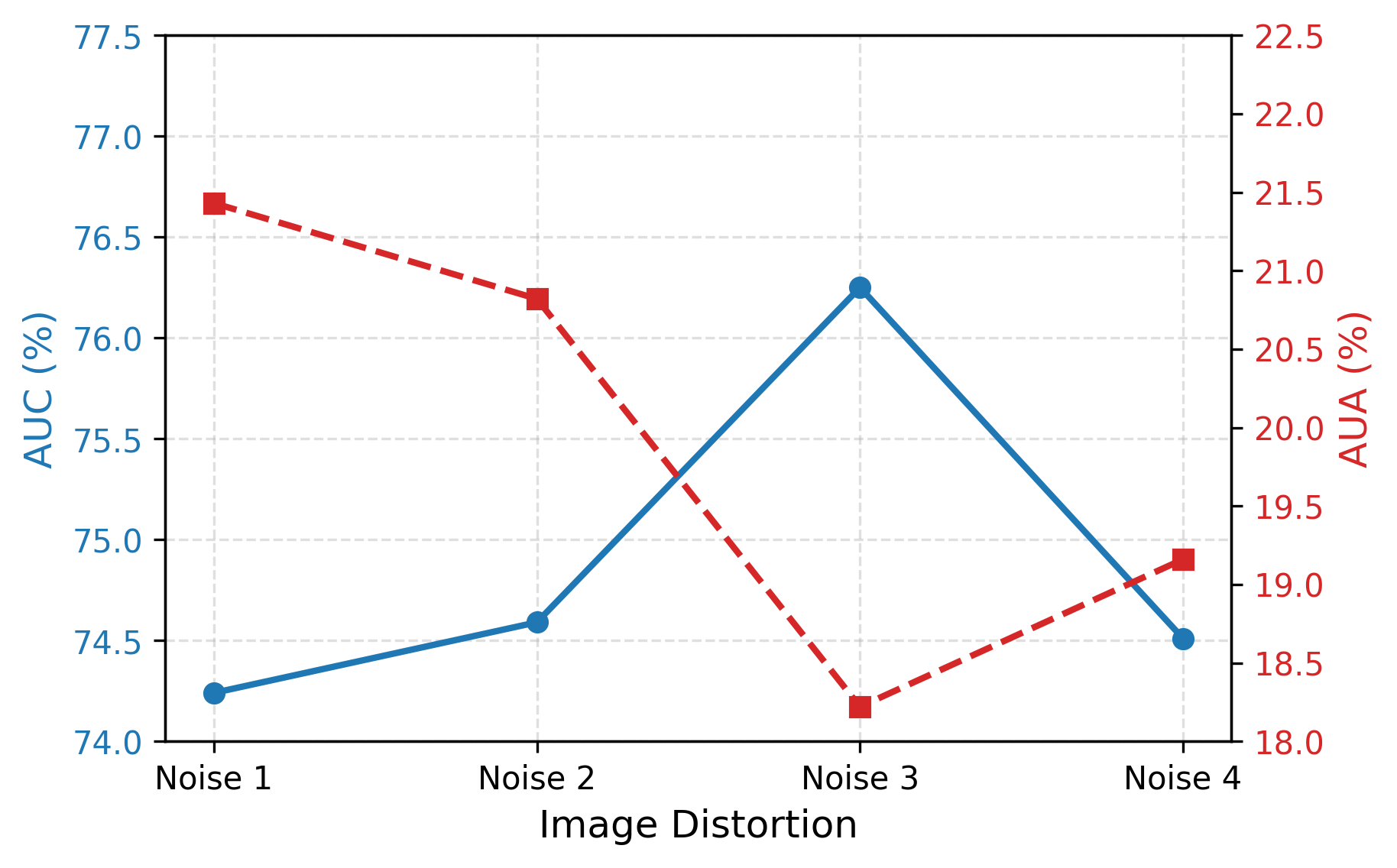}
\caption{
AUC (\%) and AUA (\%) of UniVRSE under varying image distortion intensities on the RAD-VQA dataset (MedGemma-4B-it). The solid line and dashed line correspond to AUC and AUA, respectively.
}
\label{fig:image_noise_UniVRSE}
\end{figure}

\subsection{Effect of Image Distortion Intensity}

We further investigated the impact of noise-based distortion by injecting Gaussian and Poisson noise at different intensities (Noise~1–4) on the RAD-VQA dataset using MedGemma-4B-it. In this experiment, Gaussian noise standard deviation increases from 0.03 to 0.09, paired with a Poisson scaling factor increasing from 30 to 90. 
The results in Fig.~\ref{fig:image_noise_UniVRSE} reveal a characteristic trend: moderate distortion (Noise~3) yields optimal performance (AUC 76.25\%; AUA 18.22\%), whereas both weak and excessive perturbations degrade results. This observation confirms the necessity of balancing visual diversity and semantic fidelity—mild-to-moderate distortions enhance contrastive reasoning, while overly aggressive perturbations may obscure diagnostically relevant details and impair the stability of predictive entropy. Importantly, UniVRSE maintains robustness across all noise levels, indicating that its visual contrasting mechanism generalizes well to diverse perturbation conditions.

\begin{table}[t]
\caption{
ALFA scores (\%) of three medical VLMs across four VQA and two VRG benchmarks.
Best results are highlighted in \textbf{bold}.
}
\label{tab:alfa_eval}
\centering
\setlength\tabcolsep{12pt}
\begin{tabular}{c|c|c|c}
\hline \hline
VLMs                   & $\boldsymbol{\alpha}_{\textbf{m}}$ $\uparrow$ & $\boldsymbol{\alpha}_{\textbf{h}}$ $\downarrow$ & $\boldsymbol{\alpha}_{\textbf{e}}$ $\downarrow$\\ \hline 
\multicolumn{4}{c}{RAD-VQA}  \\ \hline
MedGemma-4b-it          &	45.56	    &	35.26	    &	17.68	\\ 
LLavamed-7B        		&	27.81	    &	52.64	    &	18.05	\\ 
HuatuoGPT-Vision-7B     &	\textbf{52.50}	 &	\textbf{32.98}	&	\textbf{14.52}	\\ \hline
\multicolumn{4}{c}{SLAKE}  \\ \hline
MedGemma-4b-it          &	\textbf{54.34}	&	\textbf{24.52}	&	21.14	\\ 
LLavamed-7B             &	25.09	  &	52.90  &	21.74	\\ 
HuatuoGPT-Vision-7B     &	40.11	  &	45.01  &	\textbf{14.76}	\\ \hline
\multicolumn{4}{c}{Path-VQA}  \\ \hline
MedGemma-4b-it          &	\textbf{9.19}	&	\textbf{40.08}	 &	49.95	\\ 
LLavamed-7B             &	5.63	&	60.72	&	31.94	\\ 
HuatuoGPT-Vision-7B     &	7.65	&	60.69	&	\textbf{31.62}	\\ \hline
\multicolumn{4}{c}{MIMIC-Diff-VQA}  \\ \hline
MedGemma-4b-it          &	\textbf{18.69}	&	\textbf{38.27}	&	40.32	\\ 
LLavamed-7B             &	8.81	&	67.88	&	\textbf{15.73}	\\ 
HuatuoGPT-Vision-7B     &	12.92	&	65.46	&	17.38	\\ \hline
\multicolumn{4}{c}{ChexpertPlus}  \\ \hline
MedGemma-4b-it	        &	\textbf{25.23}	&	\textbf{25.39}	&	49.63	\\ 
LLavamed-7B	            &	3.97	&	75.16	&	\textbf{20.87}	\\ 
HuatuoGPT-Vision-7B	    &	23.94	 &	30.13	&	51.41	\\ \hline
\multicolumn{4}{c}{IU-Xray}  \\ \hline
MedGemma-4b-it 		    &	\textbf{46.26}	&	\textbf{9.55}	&	44.23	\\ 
LLavamed-7B 		    &	2.64	 &	71.57	&	\textbf{25.76}	\\ 
HuatuoGPT-Vision-7B     &	34.48	 &	21.28	 &	50.43	\\ \hline \hline
\end{tabular}
\end{table}

\begin{figure*}[t]
\centering
\includegraphics[width=\textwidth]{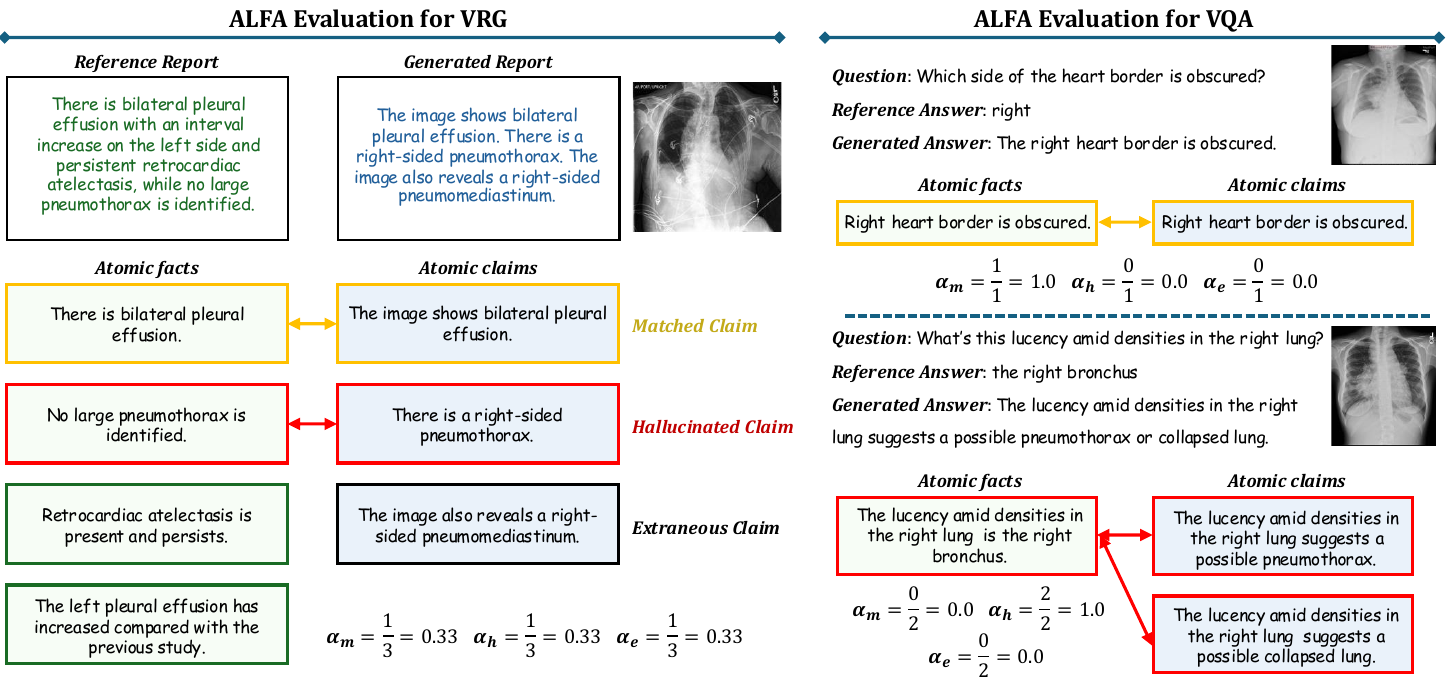}
\caption{
Examples of ALFA evaluation in medical VRG and VQA tasks. Both generated responses and reference answers are decomposed into atomic claims/facts. Semantic alignment identifies matched, hallucinated, and extraneous elements, enabling fine-grained factual consistency assessment.
}
\label{fig:image_ALFA_exam}
\end{figure*}

\subsection{ALFA Evaluation on VQA/VRG}
\label{sec:alfa_eval}
The ALFA metric was employed to evaluate the factual correctness of model-generated answers in open-ended text generation tasks. Fig.~\ref{fig:image_ALFA_exam} illustrates several examples of ALFA evaluations applied to medical VQA and VRG tasks. 
In Table~\ref{tab:alfa_eval}, we assessed the ALFA performance of three representative medical VLMs on six VQA and VRG datasets.

Overall, the results demonstrate that all three evaluated VLMs have substantial room for improvement in factual correctness across both open-ended VQA and report generation settings. A general pattern observed across datasets is the better performance on radiology-related benchmarks (RAD-VQA, SLAKE) compared to pathology-oriented ones (Path-VQA). This disparity likely reflects an imbalance in training data, which typically contains a preponderance of radiology images and reports.
For VQA tasks, MedGemma-4B-it often achieves the highest $\boldsymbol{\alpha}_{\textbf{m}}$ and lowest $\boldsymbol{\alpha}_{\textbf{h}}$, suggesting strong factual grounding in its answers. 
However, its relatively high $\boldsymbol{\alpha}_{\textbf{e}}$ scores (e.g., 49.95\% on Path-VQA and 40.32\% on MIMIC-Diff-VQA) indicate a notable tendency to include irrelevant or peripheral information. 
In contrast, HuatuoGPT-Vision-7B exhibits slightly lower $\boldsymbol{\alpha}_{\textbf{m}}$ but consistently lower $\boldsymbol{\alpha}_{\textbf{e}}$. This implies that its responses are generally more concise and better aligned with the scope of the questions. 
Conversely, LLaVAMed-7B exhibits weak factual consistency, with the highest hallucination rates on most VQA datasets.

In the more challenging VRG setting, the performance of all VLMs notably degrades. We observe universally high $\boldsymbol{\alpha}_{\textbf{e}}$ scores, often exceeding 40\%, which indicates that a large portion of the generated report content is irrelevant to the reference reports. Moreover, among the sentences judged as relevant, nearly half still contain factual errors, as reflected by significantly elevated $\boldsymbol{\alpha}_{\textbf{h}}$ scores. 

These findings highlight that while ALFA effectively quantifies the factual correctness and specific hallucination patterns of medical VLMs, achieving clinically reliable report generation remains a significant challenge. Future progress will likely depend on improving visual-grounded reasoning and integrating factual constraints during model training.

\subsection{Cross-Modality Generalization of ALFA}
\label{sec:alfa_cross_modal}

To examine the cross-modality generalization ability of ALFA, we conducted a comparative evaluation against the GREEN score~\cite{ostmeier2024green}, a specialized factuality assessment model developed for the radiology domain.  
While GREEN performs well within radiology, its evaluation reliability significantly declines when applied to datasets featuring distinct imaging modalities or clinical semantics. In sharp contrast, ALFA evaluates factuality solely through the semantic alignment of atomic facts, enabling seamless adaptation across diverse tasks and modalities.
We performed a comparative study using 100 randomly sampled test instances from both the RAD-VQA (radiology) and Path-VQA (pathology) datasets. Answers were generated by HuaTuoGPT-Vision-7B and subsequently assessed independently by both GREEN and ALFA. The accuracy of each metric was determined by manual verification of the evaluation results.
As shown in Table~\ref{tab:green_alfa}, both methods achieve comparable accuracy on the radiology dataset. However, the generalization capability of GREEN proved limited, as its performance dropped substantially from 93\% to 76\% upon transfer to the pathology domain. In stark contrast, ALFA maintained consistently high accuracy across both datasets (99\% and 93\%, respectively), thereby demonstrating superior modality generalization.

This crucial advantage primarily stems from ALFA’s task decomposition strategy. ALFA effectively reformulates the free-form factuality evaluation into two subtasks at which powerful LLMs inherently excel: atomic fact/claim decomposition and semantic equivalence judgment. By strategically leveraging the established reasoning capabilities of large language models (e.g., DeepSeek, GPT-4o), ALFA is reliably applicable to factual consistency assessment across a wide variety of report generation tasks and imaging modalities.

\begin{table}[t]
\caption{
Comparison of GREEN and ALFA evaluation accuracy (\%) on subsets of the RAD-VQA and Path-VQA datasets. Each subset contains 100 randomly sampled instances, with evaluation correctness confirmed through manual inspection.
}
\label{tab:green_alfa}
\centering
\setlength\tabcolsep{17pt}
\begin{tabular}{c|c|c}
\hline \hline
Dataset & RAD-VQA & Path-VQA \\ \hline
Modality & Radiology & Pathology \\ \hline
Acc. of Green Eval & 93 & 76 \\ 
Acc. of ALFA Eval & 99 & 93 \\ \hline \hline
\end{tabular}
\end{table}

\section{Conclusion}

This paper introduces \textbf{UniVRSE}, a vision-conditioned semantic entropy framework for detecting hallucinations in medical VLMs. UniVRSE addresses VLM overconfidence by leveraging visual contrasting, comparing semantic predictive distributions between original and distorted image-text pairs to amplify the influence of visual evidence. Complementarily, we introduced ALFA, a fine-grained metric that quantifies factual consistency at the atomic fact level, offering reliable, cross-modality hallucination supervision. Comprehensive experiments across six medical VRG/VQA datasets and three medical VLMs confirm that UniVRSE consistently outperforms prior detection methods and demonstrates robust generalization across different modalities and domains.

\bibliographystyle{IEEEtran}
\bibliography{IEEEabrv,reference}

\begin{thebibliography}{10}
\providecommand{\url}[1]{#1}
\csname url@samestyle\endcsname
\providecommand{\newblock}{\relax}
\providecommand{\bibinfo}[2]{#2}
\providecommand{\BIBentrySTDinterwordspacing}{\spaceskip=0pt\relax}
\providecommand{\BIBentryALTinterwordstretchfactor}{4}
\providecommand{\BIBentryALTinterwordspacing}{\spaceskip=\fontdimen2\font plus
\BIBentryALTinterwordstretchfactor\fontdimen3\font minus \fontdimen4\font\relax}
\providecommand{\BIBforeignlanguage}[2]{{%
\expandafter\ifx\csname l@#1\endcsname\relax
\typeout{** WARNING: IEEEtran.bst: No hyphenation pattern has been}%
\typeout{** loaded for the language `#1'. Using the pattern for}%
\typeout{** the default language instead.}%
\else
\language=\csname l@#1\endcsname
\fi
#2}}
\providecommand{\BIBdecl}{\relax}
\BIBdecl

\bibitem{xiao2024medvlmsurvey}
H.~Xiao, F.~Zhou, X.~Liu, T.~Liu, Z.~Li, X.~Liu, and X.~Huang, ``A comprehensive survey of large language models and multimodal large language models in medicine,'' \emph{arXiv preprint arXiv:2405.08603}, 2024.

\bibitem{hartsock2024vision}
I.~Hartsock and G.~Rasool, ``Vision-language models for medical report generation and visual question answering: A review,'' \emph{Frontiers in artificial intelligence}, vol.~7, p. 1430984, 2024.

\bibitem{liu2024gemex}
B.~Liu, K.~Zou, L.~Zhan, Z.~Lu, X.~Dong, Y.~Chen, C.~Xie, J.~Cao, X.-M. Wu, and H.~Fu, ``G{EM}e{X}: A large-scale, groundable, and explainable medical vqa benchmark for chest x-ray diagnosis,'' \emph{arXiv preprint arXiv:2411.16778}, 2024.

\bibitem{messina2022survey}
P.~Messina, P.~Pino, D.~Parra, A.~Soto, C.~Besa, S.~Uribe, M.~And{\'\i}a, C.~Tejos, C.~Prieto, and D.~Capurro, ``A survey on deep learning and explainability for automatic report generation from medical images,'' \emph{ACM Computing Surveys (CSUR)}, vol.~54, no. 10s, pp. 1--40, 2022.

\bibitem{huang2023llmhallusurvey}
L.~Huang, W.~Yu, W.~Ma, W.~Zhong, Z.~Feng, H.~Wang, Q.~Chen, W.~Peng, X.~Feng, B.~Qin \emph{et~al.}, ``A survey on hallucination in large language models: Principles, taxonomy, challenges, and open questions,'' \emph{ACM Transactions on Information Systems}, 2023.

\bibitem{liu2024mllmhallusurvey}
H.~Liu, W.~Xue, Y.~Chen, D.~Chen, X.~Zhao, K.~Wang, L.~Hou, R.~Li, and W.~Peng, ``A survey on hallucination in large vision-language models,'' \emph{arXiv preprint arXiv:2402.00253}, 2024.

\bibitem{yu2024hallucidoctor}
Q.~Yu, J.~Li, L.~Wei, L.~Pang, W.~Ye, B.~Qin, S.~Tang, Q.~Tian, and Y.~Zhuang, ``Hallucidoctor: Mitigating hallucinatory toxicity in visual instruction data,'' in \emph{CVPR}, 2024, pp. 12\,944--12\,953.

\bibitem{yang2025mitigating}
Z.~Yang, X.~Luo, D.~Han, Y.~Xu, and D.~Li, ``Mitigating hallucinations in large vision-language models via dpo: On-policy data hold the key,'' in \emph{Proceedings of the Computer Vision and Pattern Recognition Conference}, 2025, pp. 10\,610--10\,620.

\bibitem{jiang2024hallucination}
C.~Jiang, H.~Xu, M.~Dong, J.~Chen, W.~Ye, M.~Yan, Q.~Ye, J.~Zhang, F.~Huang, and S.~Zhang, ``Hallucination augmented contrastive learning for multimodal large language model,'' in \emph{CVPR}, 2024, pp. 27\,036--27\,046.

\bibitem{zhuang2025vasparse}
X.~Zhuang, Z.~Zhu, Y.~Xie, L.~Liang, and Y.~Zou, ``Vasparse: Towards efficient visual hallucination mitigation via visual-aware token sparsification,'' in \emph{Proceedings of the Computer Vision and Pattern Recognition Conference}, 2025, pp. 4189--4199.

\bibitem{wang2024mitigating}
X.~Wang, J.~Pan, L.~Ding, and C.~Biemann, ``Mitigating hallucinations in large vision-language models with instruction contrastive decoding,'' in \emph{ACL}, 2024, pp. 15\,840--15\,853.

\bibitem{gunjal2024detecting}
A.~Gunjal, J.~Yin, and E.~Bas, ``Detecting and preventing hallucinations in large vision language models,'' in \emph{AAAI}, vol.~38, no.~16, 2024, pp. 18\,135--18\,143.

\bibitem{xiao2024detecting}
W.~Xiao, Z.~Huang, L.~Gan, W.~He, H.~Li, Z.~Yu, H.~Jiang, F.~Wu, and L.~Zhu, ``Detecting and mitigating hallucination in large vision language models via fine-grained ai feedback,'' \emph{arXiv preprint arXiv:2404.14233}, 2024.

\bibitem{chen2024detecting}
J.~Chen, D.~Yang, T.~Wu, Y.~Jiang, X.~Hou, M.~Li, S.~Wang, D.~Xiao, K.~Li, and L.~Zhang, ``Detecting and evaluating medical hallucinations in large vision language models,'' \emph{arXiv preprint arXiv:2406.10185}, 2024.

\bibitem{hardy2024rextrust}
R.~Hardy, S.~E. Kim, and P.~Rajpurkar, ``Rextrust: A model for fine-grained hallucination detection in ai-generated radiology reports,'' \emph{arXiv preprint arXiv:2412.15264}, 2024.

\bibitem{cohen2023lm}
R.~Cohen, M.~Hamri, M.~Geva, and A.~Globerson, ``Lm vs lm: Detecting factual errors via cross examination,'' in \emph{EMNLP}, 2023, pp. 12\,621--12\,640.

\bibitem{sahu2024pelican}
P.~Sahu, K.~Sikka, and A.~Divakaran, ``Pelican: Correcting hallucination in vision-llms via claim decomposition and program of thought verification,'' in \emph{EMNLP}, 2024, pp. 8228--8248.

\bibitem{yin2024woodpecker}
S.~Yin, C.~Fu, S.~Zhao, T.~Xu, H.~Wang, D.~Sui, Y.~Shen, K.~Li, X.~Sun, and E.~Chen, ``Woodpecker: Hallucination correction for multimodal large language models,'' \emph{Science China Information Sciences}, vol.~67, no.~12, p. 220105, 2024.

\bibitem{chen2024complex}
J.~Chen, G.~Kim, A.~Sriram, G.~Durrett, and E.~Choi, ``Complex claim verification with evidence retrieved in the wild,'' in \emph{Proceedings of the 2024 Conference of the North American Chapter of the Association for Computational Linguistics: Human Language Technologies}, 2024, pp. 3569--3587.

\bibitem{min2023factscore}
S.~Min, K.~Krishna, X.~Lyu, M.~Lewis, W.-t. Yih, P.~Koh, M.~Iyyer, L.~Zettlemoyer, and H.~Hajishirzi, ``Factscore: Fine-grained atomic evaluation of factual precision in long form text generation,'' in \emph{EMNLP}, 2023, pp. 12\,076--12\,100.

\bibitem{li2024reference}
Q.~Li, J.~Geng, C.~Lyu, D.~Zhu, M.~Panov, and F.~Karray, ``Reference-free hallucination detection for large vision-language models,'' in \emph{EMNLP}, 2024, pp. 4542--4551.

\bibitem{farquhar2024detecting}
S.~Farquhar, J.~Kossen, L.~Kuhn, and Y.~Gal, ``Detecting hallucinations in large language models using semantic entropy,'' \emph{Nature}, vol. 630, no. 8017, pp. 625--630, 2024.

\bibitem{chen2024inside}
C.~Chen, K.~Liu, Z.~Chen, Y.~Gu, Y.~Wu, M.~Tao, Z.~Fu, and J.~Ye, ``I{NSIDE}: L{LM}s' internal states retain the power of hallucination detection,'' in \emph{ICLR}, 2024.

\bibitem{zou2023review}
K.~Zou, Z.~Chen, X.~Yuan, X.~Shen, M.~Wang, and H.~Fu, ``A review of uncertainty estimation and its application in medical imaging,'' \emph{Meta-Radiology}, p. 100003, 2023.

\bibitem{zhang2024vl}
R.~Zhang, H.~Zhang, and Z.~Zheng, ``Vl-uncertainty: Detecting hallucination in large vision-language model via uncertainty estimation,'' \emph{arXiv preprint arXiv:2411.11919}, 2024.

\bibitem{sellergren2025medgemma}
A.~Sellergren, S.~Kazemzadeh, T.~Jaroensri, A.~Kiraly, M.~Traverse, T.~Kohlberger, S.~Xu, F.~Jamil, C.~Hughes, C.~Lau \emph{et~al.}, ``Medgemma technical report,'' \emph{arXiv preprint arXiv:2507.05201}, 2025.

\bibitem{li2024llavamed}
C.~Li, C.~Wong, S.~Zhang, N.~Usuyama, H.~Liu, J.~Yang, T.~Naumann, H.~Poon, and J.~Gao, ``L{L}a{VA}-{M}ed: Training a large language-and-vision assistant for biomedicine in one day,'' \emph{NeurIPS}, vol.~36, 2024.

\bibitem{chen2024towards}
J.~Chen, C.~Gui, R.~Ouyang, A.~Gao, S.~Chen, G.~H. Chen, X.~Wang, Z.~Cai, K.~Ji, X.~Wan \emph{et~al.}, ``Towards injecting medical visual knowledge into multimodal llms at scale,'' in \emph{Proceedings of the 2024 conference on empirical methods in natural language processing}, 2024, pp. 7346--7370.

\bibitem{han2024semantic}
J.~Han, J.~Kossen, M.~Razzak, L.~Schut, S.~A. Malik, and Y.~Gal, ``Semantic entropy probes: Robust and cheap hallucination detection in llms,'' in \emph{ICML Workshop on Foundation Models in the Wild}, 2024.

\bibitem{zhang2023sac}
J.~Zhang, Z.~Li, K.~Das, B.~A. Malin, and S.~Kumar, ``Sac3: Reliable hallucination detection in black-box language models via semantic-aware cross-check consistency,'' in \emph{The 2023 Conference on Empirical Methods in Natural Language Processing}, 2023.

\bibitem{varshney2023stitch}
N.~Varshney, W.~Yao, H.~Zhang, J.~Chen, and D.~Yu, ``A stitch in time saves nine: Detecting and mitigating hallucinations of llms by validating low-confidence generation,'' \emph{arXiv preprint arXiv:2307.03987}, 2023.

\bibitem{sun2025redeep}
Z.~Sun, X.~Zang, K.~Zheng, J.~Xu, X.~Zhang, W.~Yu, Y.~Song, and H.~Li, ``Redeep: Detecting hallucination in retrieval-augmented generation via mechanistic interpretability,'' in \emph{The Thirteenth International Conference on Learning Representations}, 2025.

\bibitem{kuhnsemantic}
L.~Kuhn, Y.~Gal, and S.~Farquhar, ``Semantic uncertainty: Linguistic invariances for uncertainty estimation in natural language generation,'' in \emph{The Eleventh International Conference on Learning Representations}, 2023.

\bibitem{manakul2023selfcheckgpt}
P.~Manakul, A.~Liusie, and M.~Gales, ``Selfcheckgpt: Zero-resource black-box hallucination detection for generative large language models,'' in \emph{EMNLP}, 2023, pp. 9004--9017.

\bibitem{agrawal2024language}
A.~Agrawal, M.~Suzgun, L.~Mackey, and A.~Kalai, ``Do language models know when they’re hallucinating references?'' in \emph{Findings of the Association for Computational Linguistics: EACL 2024}, 2024, pp. 912--928.

\bibitem{whitehead2024pre}
S.~Whitehead, J.~Phillips, and S.~M. Hendryx, ``Pre-training multimodal hallucination detectors with corrupted grounding data,'' in \emph{Neurips Safe Generative AI Workshop}, 2024.

\bibitem{wu2024logical}
J.~Wu, Q.~Liu, D.~Wang, J.~Zhang, S.~Wu, L.~Wang, and T.~Tan, ``Logical closed loop: Uncovering object hallucinations in large vision-language models,'' in \emph{Findings of the Association for Computational Linguistics ACL 2024}, 2024, pp. 6944--6962.

\bibitem{zhang2024dhcp}
Y.~Zhang, R.~Xie, J.~Chen, X.~Sun, Y.~Wang \emph{et~al.}, ``Dhcp: Detecting hallucinations by cross-modal attention pattern in large vision-language models,'' \emph{arXiv preprint arXiv:2411.18659}, 2024.

\bibitem{sambara2024radflag}
S.~Zhang, S.~Sambara, O.~Banerjee, J.~N. Acosta, L.~J. Fahrner, and P.~Rajpurkar, ``Radflag: A black-box hallucination detection method for medical vision language models,'' in \emph{Machine Learning for Health}.\hskip 1em plus 0.5em minus 0.4em\relax PMLR, 2025, pp. 1087--1103.

\bibitem{chen2024unified}
X.~Chen, C.~Wang, N.~Zhang, Y.~Xue, Y.~SHEN, G.~Jinjie, H.~Chen \emph{et~al.}, ``Unified hallucination detection for multimodal large language models,'' in \emph{ICLR Workshop on Reliable and Responsible Foundation Models}, 2024.

\bibitem{kim2024esreal}
M.~Kim, M.~Kim, J.~Bae, S.~Choi, S.~Kim, and B.~Chang, ``Esreal: Exploiting semantic reconstruction to mitigate hallucinations in vision-language models,'' in \emph{European Computer Vision Association}, 2024.

\bibitem{park2025convis}
Y.~Park, D.~Lee, J.~Choe, and B.~Chang, ``Convis: Contrastive decoding with hallucination visualization for mitigating hallucinations in multimodal large language models,'' in \emph{Proceedings of the AAAI Conference on Artificial Intelligence}, vol.~39, no.~6, 2025, pp. 6434--6442.

\bibitem{zhang2024meter}
R.~Zhang, J.~Chen, M.~Dai, X.~Jiang, Y.~Hu, B.~Liu, and J.~Cao, ``Meter: Multimodal hallucination detection with mixture of experts via tools ensembling and reasoning,'' in \emph{CCF International Conference on Natural Language Processing and Chinese Computing}.\hskip 1em plus 0.5em minus 0.4em\relax Springer, 2024, pp. 274--286.

\bibitem{liao2025VASE}
Z.~Liao, S.~Hu, K.~Zou, H.~Fu, L.~Zhen, and Y.~Xia, ``Vision-amplified semantic entropy for hallucination detection in medical visual question answering,'' in \emph{International conference on medical image computing and computer-assisted intervention}.\hskip 1em plus 0.5em minus 0.4em\relax Springer, 2025.

\bibitem{papineni2002bleu}
K.~Papineni, S.~Roukos, T.~Ward, and W.-J. Zhu, ``Bleu: a method for automatic evaluation of machine translation,'' in \emph{Proceedings of the 40th annual meeting of the Association for Computational Linguistics}, 2002, pp. 311--318.

\bibitem{lin2004rouge}
C.-Y. Lin, ``Rouge: A package for automatic evaluation of summaries,'' in \emph{Text summarization branches out}, 2004, pp. 74--81.

\bibitem{banerjee2005meteor}
S.~Banerjee and A.~Lavie, ``Meteor: An automatic metric for mt evaluation with improved correlation with human judgments,'' in \emph{Proceedings of the acl workshop on intrinsic and extrinsic evaluation measures for machine translation and/or summarization}, 2005, pp. 65--72.

\bibitem{zhang2019bertscore}
T.~Zhang, V.~Kishore, F.~Wu, K.~Q. Weinberger, and Y.~Artzi, ``Bertscore: Evaluating text generation with bert,'' in \emph{International Conference on Learning Representations}, 2020.

\bibitem{chang2025medheval}
A.~Chang, L.~Huang, P.~Bhatia, T.~Kass-Hout, F.~Ma, and C.~Xiao, ``Medheval: Benchmarking hallucinations and mitigation strategies in medical large vision-language models,'' \emph{arXiv preprint arXiv:2503.02157}, 2025.

\bibitem{zuo2025medhallbench}
K.~Zuo and Y.~Jiang, ``Medhallbench: A new benchmark for assessing hallucination in medical large language models,'' in \emph{AAAI Bridge Program on AI for Medicine and Healthcare}.\hskip 1em plus 0.5em minus 0.4em\relax PMLR, 2025, pp. 205--213.

\bibitem{yan2025medhalltune}
Q.~Yan, Y.~Yuan, X.~Hu, Y.~Wang, J.~Xu, J.~Li, C.-W. Fu, and P.-A. Heng, ``Medhalltune: An instruction-tuning benchmark for mitigating medical hallucination in vision-language models,'' \emph{arXiv preprint arXiv:2502.20780}, 2025.

\bibitem{ostmeier2024green}
S.~Ostmeier, J.~Xu, Z.~Chen, M.~Varma, L.~Blankemeier, C.~Bluethgen, A.~E. Michalson, M.~Moseley, C.~Langlotz, A.~S. Chaudhari \emph{et~al.}, ``Green: Generative radiology report evaluation and error notation,'' \emph{arXiv preprint arXiv:2405.03595}, 2024.

\bibitem{liu2024deepseek}
A.~Liu, B.~Feng, B.~Xue, B.~Wang, B.~Wu, C.~Lu, C.~Zhao, C.~Deng, C.~Zhang, C.~Ruan \emph{et~al.}, ``Deepseek-v3 technical report,'' \emph{arXiv preprint arXiv:2412.19437}, 2024.

\bibitem{he2020deberta}
P.~He, X.~Liu, J.~Gao, and W.~Chen, ``Deberta: Decoding-enhanced bert with disentangled attention,'' in \emph{ICLR}, 2020.

\bibitem{peng2025enhancing}
Y.~Peng, A.~Lin, M.~Wang, T.~Lin, L.~Liu, J.~Wu, K.~Zou, T.~Shi, L.~Feng, Z.~Liang \emph{et~al.}, ``Enhancing ai reliability: A foundation model with uncertainty estimation for optical coherence tomography-based retinal disease diagnosis,'' \emph{Cell Reports Medicine}, vol.~6, no.~1, 2025.

\bibitem{hurst2024gpt}
A.~Hurst, A.~Lerer, A.~P. Goucher, A.~Perelman, A.~Ramesh, A.~Clark, A.~Ostrow, A.~Welihinda, A.~Hayes, A.~Radford \emph{et~al.}, ``Gpt-4o system card,'' \emph{arXiv preprint arXiv:2410.21276}, 2024.

\bibitem{lau2018dataset}
J.~J. Lau, S.~Gayen, A.~Ben~Abacha, and D.~Demner-Fushman, ``A dataset of clinically generated visual questions and answers about radiology images,'' \emph{Scientific data}, vol.~5, no.~1, pp. 1--10, 2018.

\bibitem{liu2021slake}
B.~Liu, L.-M. Zhan, L.~Xu, L.~Ma, Y.~Yang, and X.-M. Wu, ``Slake: A semantically-labeled knowledge-enhanced dataset for medical visual question answering,'' in \emph{2021 IEEE 18th international symposium on biomedical imaging (ISBI)}.\hskip 1em plus 0.5em minus 0.4em\relax IEEE, 2021, pp. 1650--1654.

\bibitem{he2020pathvqa}
X.~He, Y.~Zhang, L.~Mou, E.~Xing, and P.~Xie, ``Pathvqa: 30000+ questions for medical visual question answering,'' \emph{arXiv preprint arXiv:2003.10286}, 2020.

\bibitem{hu2023expert}
X.~Hu, L.~Gu, Q.~An, M.~Zhang, L.~Liu, K.~Kobayashi, T.~Harada, R.~M. Summers, and Y.~Zhu, ``Expert knowledge-aware image difference graph representation learning for difference-aware medical visual question answering,'' in \emph{Proceedings of the 29th ACM SIGKDD Conference on Knowledge Discovery and Data Mining}, 2023, pp. 4156--4165.

\bibitem{chambon2024chexpertplus}
P.~Chambon, J.-B. Delbrouck, T.~Sounack, S.-C. Huang, Z.~Chen, M.~Varma, S.~Q. Truong, C.~T. Chuong, and C.~P. Langlotz, ``Chexpert plus: Augmenting a large chest x-ray dataset with text radiology reports, patient demographics and additional image formats,'' \emph{arXiv preprint arXiv:2405.19538}, 2024.

\bibitem{demner2015preparing}
D.~Demner-Fushman, M.~D. Kohli, M.~B. Rosenman, S.~E. Shooshan, L.~Rodriguez, S.~Antani, G.~R. Thoma, and C.~J. McDonald, ``Preparing a collection of radiology examinations for distribution and retrieval,'' \emph{Journal of the American Medical Informatics Association}, vol.~23, no.~2, pp. 304--310, 2015.

\bibitem{johnson2019mimic}
A.~E. Johnson, T.~J. Pollard, S.~J. Berkowitz, N.~R. Greenbaum, M.~P. Lungren, C.-y. Deng, R.~G. Mark, and S.~Horng, ``M{IMIC}-{CXR}, a de-identified publicly available database of chest radiographs with free-text reports,'' \emph{Scientific data}, vol.~6, no.~1, p. 317, 2019.

\end{thebibliography}


\vspace{-25pt}
\begin{IEEEbiography}[{\includegraphics[width=1in,height=1.25in,clip,keepaspectratio]{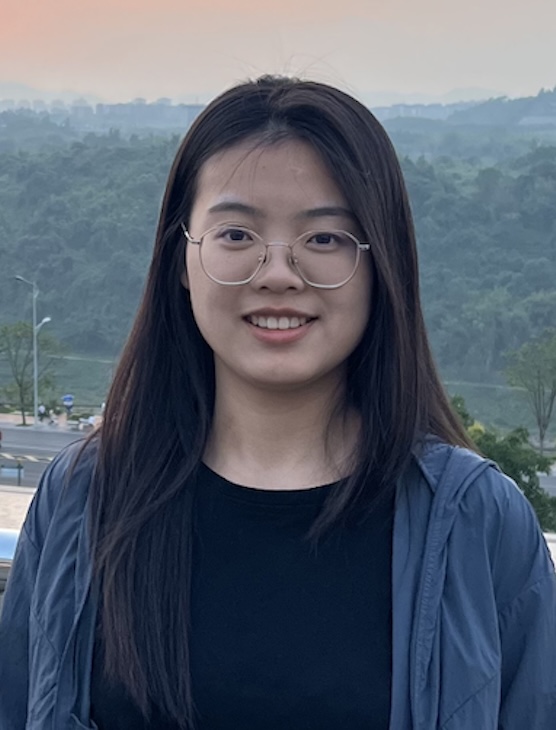}}]{Zehui Liao} 
is currently a Ph.D. candidate in the School of Computer Science at Northwestern Polytechnical University, supervised by Prof. Yong Xia.
From 2024 to 2025, she was a visiting Ph.D. student at the Institute of High Performance Computing, Agency for Science, Technology and Research, Singapore, under the supervision of Dr. Huazhu Fu and Dr. Liangli Zhen.
Her research interests include medical image analysis, with a particular focus on hallucination in medical vision–language models and learning under label uncertainty.
\end{IEEEbiography}

\vspace{-25pt}
\begin{IEEEbiography}[{\includegraphics[width=1in,height=1.25in,clip,keepaspectratio]{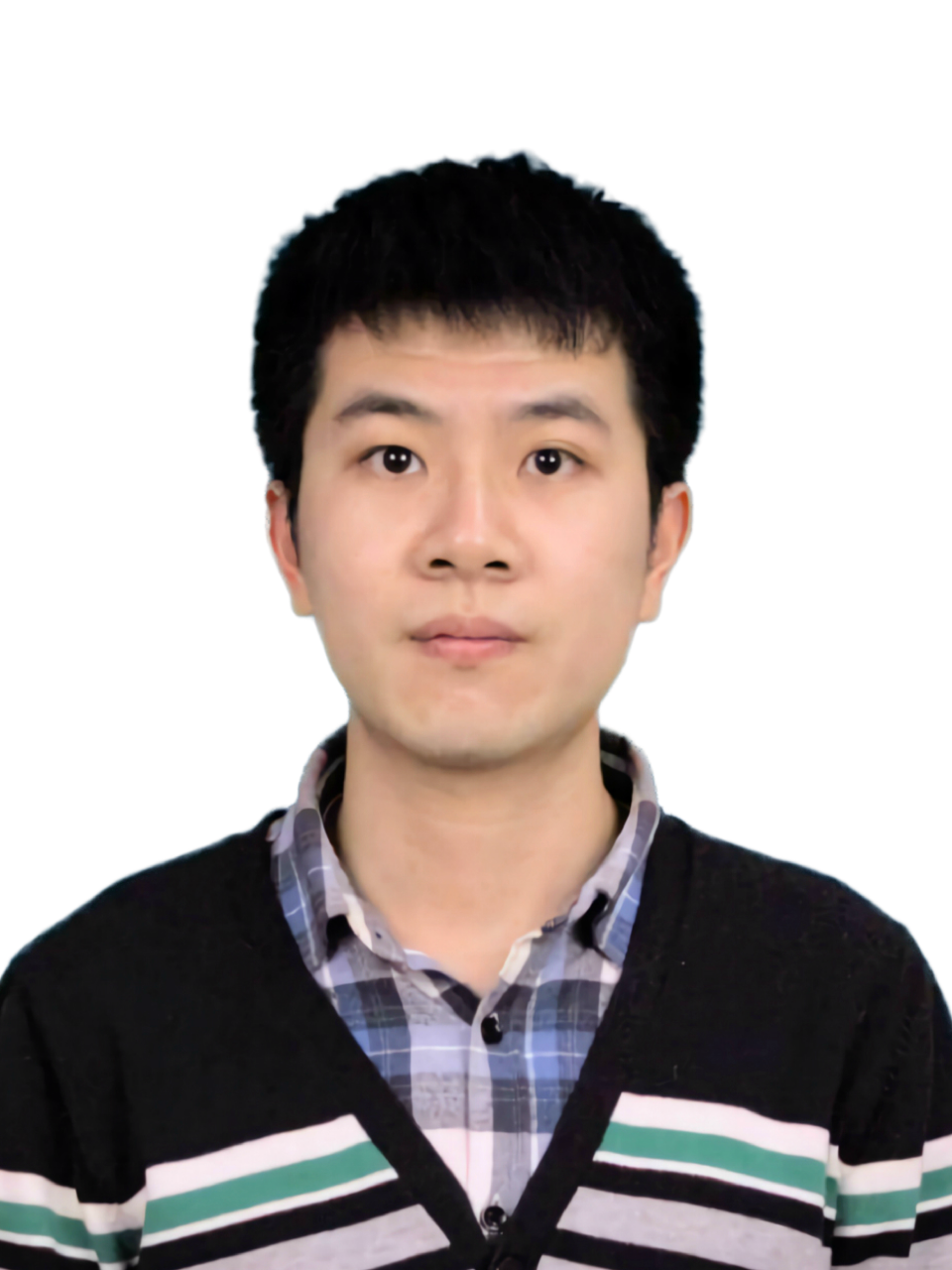}}]{Shishuai Hu} 
is currently a Ph.D. candidate in the School of Computer Science at Northwestern Polytechnical University, supervised by Prof. Yong Xia.
From 2024 to 2025, He was a visiting Ph.D. student at the Institute of High Performance Computing, Agency for Science, Technology and Research, Singapore, under the supervision of Dr. Huazhu Fu and Dr. Liangli Zhen.
His research interests include medical image analysis and AI for healthcare, with a particular focus on medical image segmentation, domain adaptation, and in-context learning.
\end{IEEEbiography}

\begin{IEEEbiography}[{\includegraphics[width=1in,height=1.25in,clip,keepaspectratio]{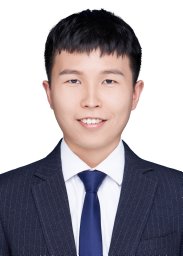}}]{Ke Zou} is currently a research fellow at the National University of Singapore. He received his PhD degree in 2024 from the National Key Laboratory of Fundamental Science on Synthetic Vision at the College of Computer Science, Sichuan University, Chengdu, China. From 2022 to 2023, he served as a visiting PhD student at the Institute of High Performance Computing, Agency for Science, Technology and Research in Singapore. His research interests encompass medical image analysis, vision-language models, and uncertainty estimation. 
\end{IEEEbiography}

\vspace{-20pt}
\begin{IEEEbiography}[{\includegraphics[width=1in,height=1.25in,clip,keepaspectratio]{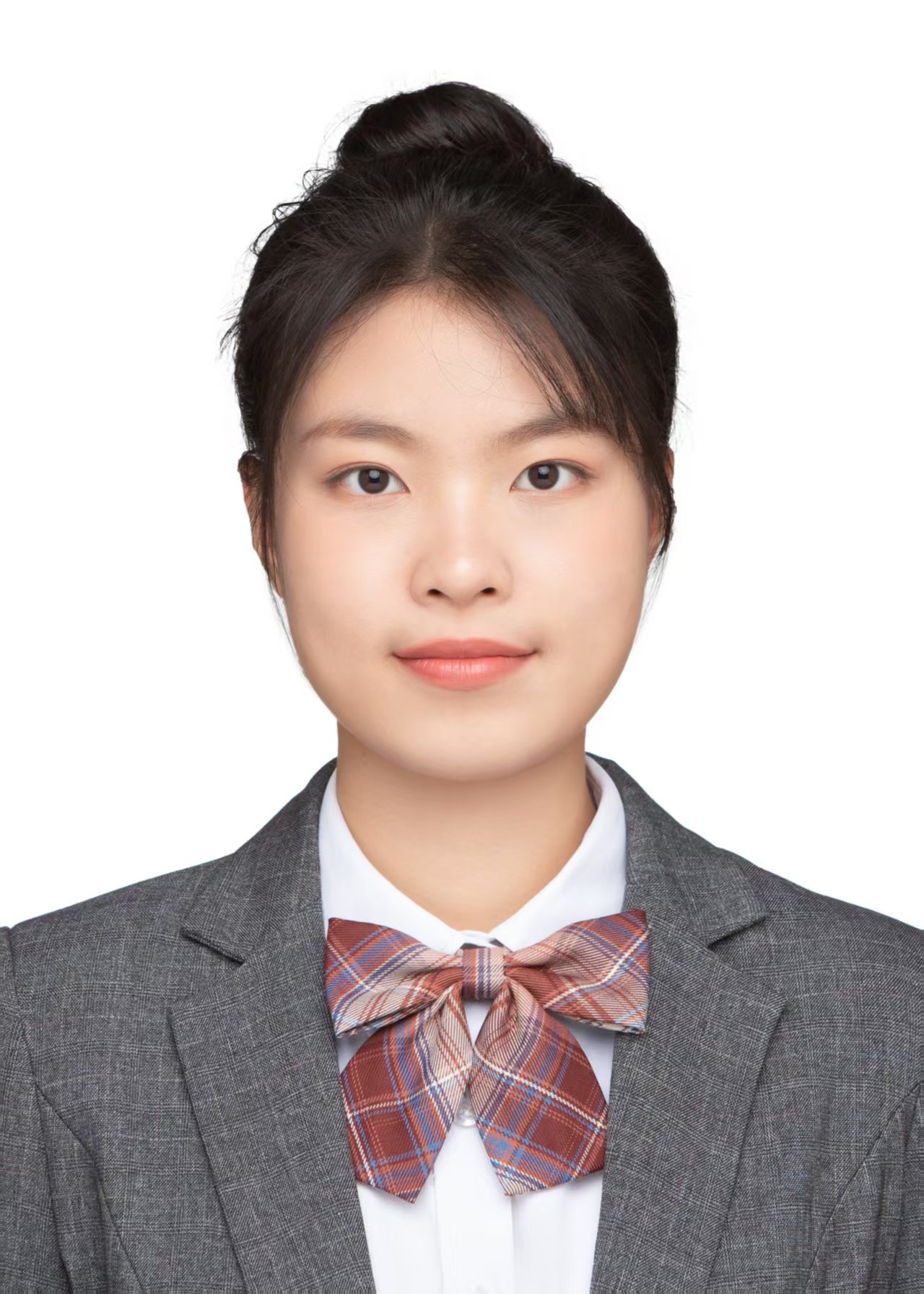}}]{Mengyuan Jin} received the B.E. degree from the School of Computer Science, Northwestern Polytechnical University (NPU), Xi’an, China, in 2024.
She is currently a PhD candidate in the School of Computer Science at NPU, supervised by Prof. Yong Xia. 
Her research focuses on medical image analysis, hallucination detection and mitigation, and medical vision-language models.
\end{IEEEbiography}

\vspace{-20pt}
\begin{IEEEbiography}[{\includegraphics[width=1in,height=1.25in,clip,keepaspectratio]{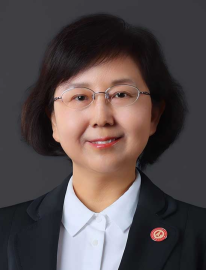}}]{Yanning Zhang} (Fellow, IEEE) 
received the B.S. degree from the Department of Electronic Engineering, Dalian University of Technology, Dalian, China, in 1988, the M.S. degree from the School of Electronic Engineering, and the Ph.D. degree from the School of Marine Engineering, Northwestern Polytechnical University (NPU), Xian, China, in 1993 and 1996, respectively. She is currently a professor with the School of Computer Science at NPU. Her research interests include computer vision and pattern recognition, image and video processing, and intelligent information processing. 
\end{IEEEbiography}

\vspace{-22pt}
\begin{IEEEbiography}[{\includegraphics[width=1in,height=1.25in,clip,keepaspectratio]{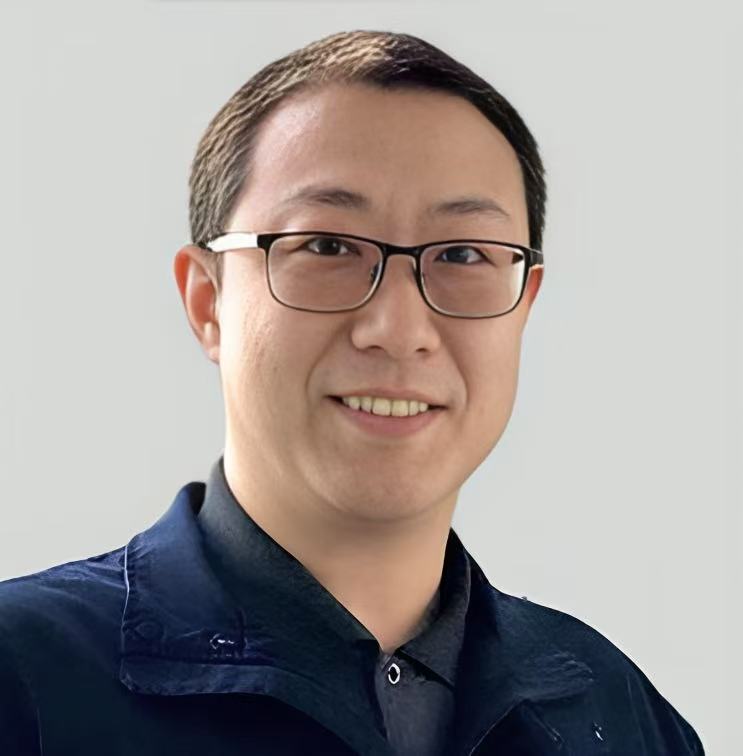}}]{Huazhu Fu}
(Senior Member, IEEE) is a Principal Scientist at the Institute of High Performance Computing (IHPC), A*STAR, Singapore. He earned his Ph.D. from Tianjin University in 2013. His research focuses on medical image analysis, AI for healthcare, and trustworthy AI. He is an Associate Editor for several distinguished journals, including IEEE TMI, IEEE TNNLS, and IEEE JBHI.  
\end{IEEEbiography}

\vspace{-22pt}
\begin{IEEEbiography}[{\includegraphics[width=1in,height=1.25in,clip,keepaspectratio]{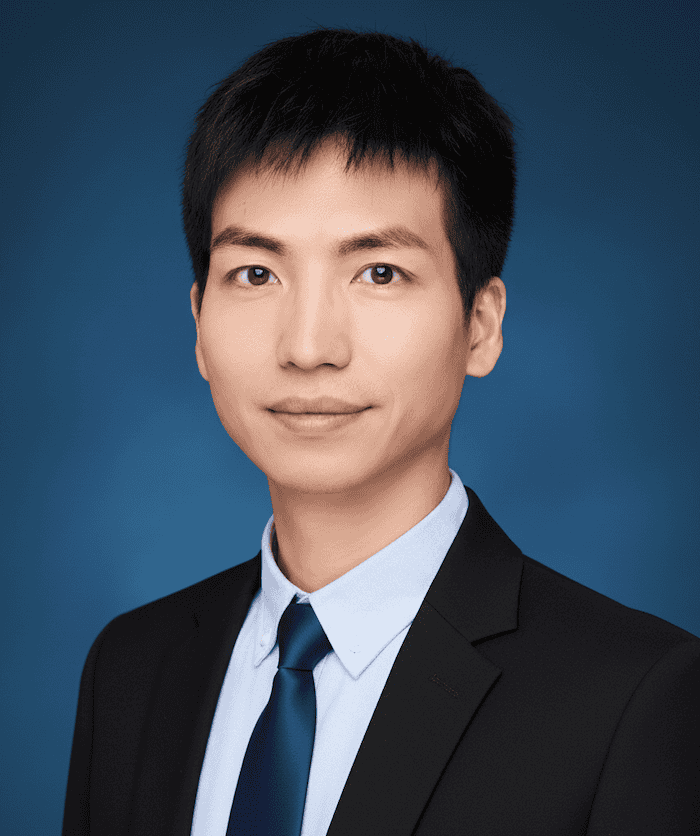}}]{Liangli Zhen} (Senior Member, IEEE) is a senior scientist at the Institute of High Performance Computing, Agency for Science, Technology and Research (A*STAR), Singapore. 
Prior to joining A*STAR, he received his PhD in Computer Science from Sichuan University, China, in 2018. From 2016 to 2018, he was a joint PhD student at The University of Birmingham in the UK. His research is mainly on machine learning and optimisation, especially AI safety, multimodal learning, and multi-objective optimisation.
\end{IEEEbiography}

\vspace{-20pt}
\begin{IEEEbiography}[{\includegraphics[width=1in,height=1.25in,clip,keepaspectratio]{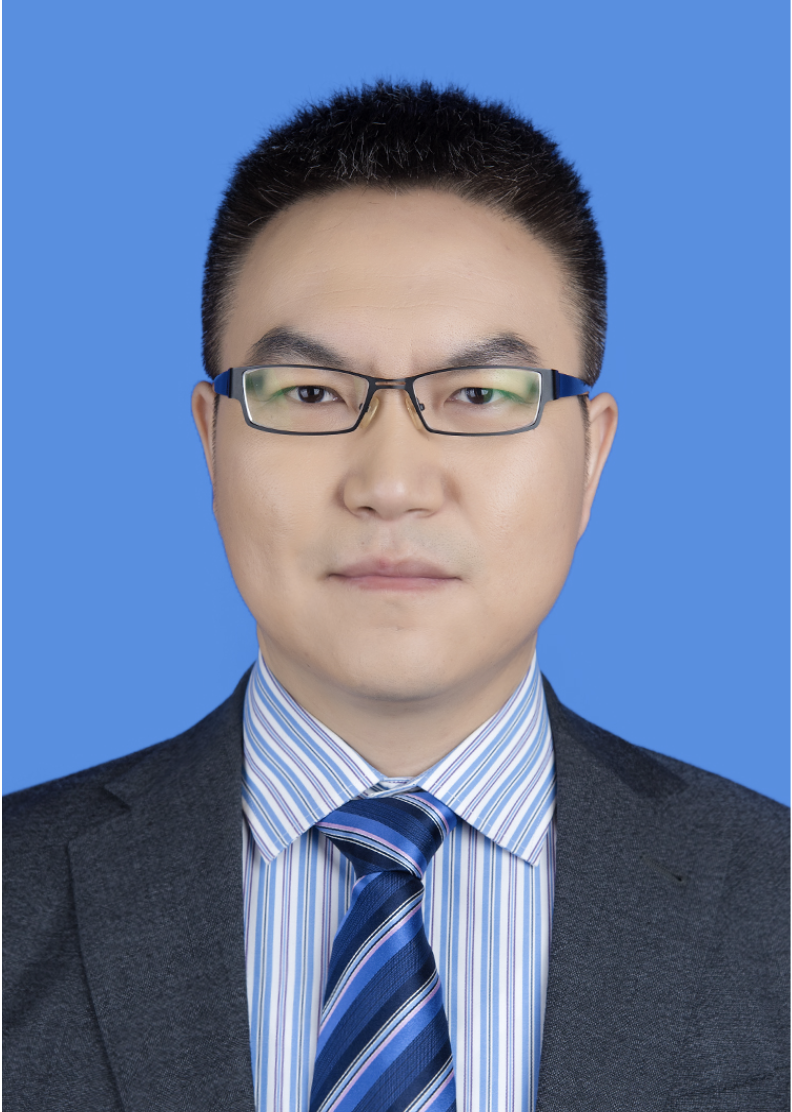}}]{Yong Xia} (Senior Member, IEEE) received the PhD degree in computer science and technology from Northwestern Polytechnical University (NPU), Xi’an, China, in 2007. He is currently a professor with the School of Computer Science and Engineering, NPU. His research interests include medical image analysis and deep learning. Over the past five years, he has jointly published more than 80 research papers in JAMA Network Open, Radiology, IEEE Transactions on Pattern Analysis and Machine Intelligence/IEEE Transactions on Medical Imaging/IEEE Transactions on Image Processing, MedIA, NeurIPS, CVPR, ECCV, AAAI, IJCAI, and MICCAI.
\end{IEEEbiography}

\vfill

\end{document}